\documentclass[10pt,twocolumn,letterpaper]{article}
\pdfoutput=1

\usepackage{wacv}
\usepackage{times}
\usepackage{epsfig}
\usepackage{graphicx}
\usepackage{amsmath}
\usepackage{amssymb}

\makeatletter
\@namedef{ver@everyshi.sty}{}
\makeatother

\usepackage[table]{xcolor}
\usepackage[nohyperlinks,nolist]{acronym}
\usepackage[binary-units=true]{siunitx}
\usepackage{multirow}
\usepackage{makecell}
\usepackage{xspace}
\usepackage[inline]{enumitem}
\usepackage{subcaption}

\def\wacvPaperID{0} % Enter the WACV Paper ID here

\wacvfinalcopy % *** Uncomment this line for the final submission

\ifwacvfinal
\def\assignedStartPage{1} % *** Enter the assigned starting page number (instead of 9876)
\fi

\usepackage{hyperref}

\ifwacvfinal
\else
\fi

\begin{document}

\begin{acronym}
   \acro{GAN}{Generative Adversarial Network}
   \acro{AdaIN}{Adaptive Instance Normalization}
   \acro{LPIPS}{Learned Perceptual Image Patch Similarity}
   \acro{FID}{Frechet Inception Distance}
   \acro{PSNR}{Peak Signal-to-Noise Ratio}
   \acro{SSIM}{Structural Similarity Index}
   \acro{MSE}{Mean Squared Error}
   \acro{CNN}{Convolutional Neural Network}
\end{acronym}

\newcommand{\fid}{\footnotesize FID$^\downarrow$}
\newcommand{\psnr}{\footnotesize PSNR$^\uparrow$}
\newcommand{\ssim}{\footnotesize SSIM$^\uparrow$}
\newcommand{\gc}{\cellcolor{blue!15}}
\newcommand{\stylegan}{StyleGAN\xspace}

\graphicspath{{./images/}}

%%%%%%%%% TITLE
% \title{One Model to Generate Them All:\\ Beyond the Latent Code in \stylegan - About the Importance of Noise}
% \title{One Model to Reconstruct Them All:\\ Beyond the Latent Code in \stylegan - About the Importance of Noise}
% \title{One Model to Reconstruct Them All:\\ A Novel Way to make use of Stochastic Noise in \stylegan}
\title{One Model to Reconstruct Them All:\\ A Novel Way to Use the Stochastic Noise in \stylegan}
% \title{One Model to Reconstruct Them All:\\ Understanding the Stochastic Noise in \stylegan}

\author{Joseph Bethge\thanks{Equal Contribution}, Christian Bartz\footnotemark[1], Haojin Yang, Christoph Meinel\\
Hasso Plattner Insititute\\
University of Potsdam\\
{\tt\small \{firstname.lastname\}@hpi.de}
}

\maketitle

\begin{abstract}
\noindent
    \acp{GAN} have achieved state-of-the-art performance for several image generation and manipulation tasks.
    Different works have improved the limited understanding of the latent space of \acp{GAN} by embedding images into specific \ac{GAN} architectures to reconstruct the original images.
    We present a novel \stylegan-based autoencoder architecture, which can reconstruct images with very high quality across several data domains.
    We demonstrate a previously unknown grade of generalizablility by training the encoder and decoder independently and on different datasets.
    Furthermore, we provide new insights about the significance and capabilities of noise inputs of the well-known \stylegan architecture.
    Our proposed architecture can handle up to 40 images per second on a single GPU, which is approximately 28$\times$ faster than previous approaches.
    Finally, our model also shows promising results, when compared to the state-of-the-art on the image denoising task, although it was not explicitly designed for this task.
\end{abstract}
\section{Introduction}

\begin{figure}
  \begin{center}
  \includegraphics[width=\linewidth]{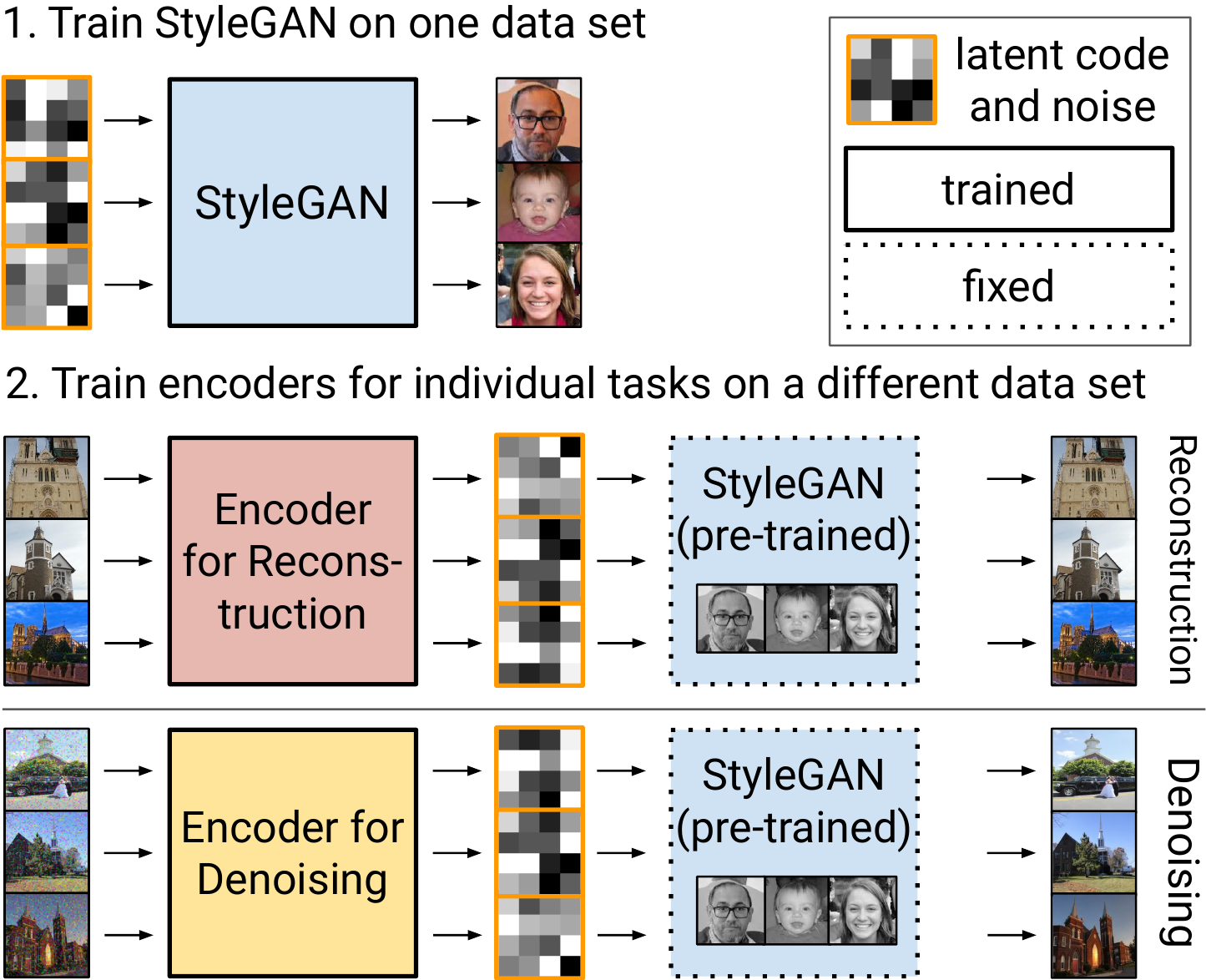}
  \end{center}
  \vspace{-0.2cm}
  \caption{
  Our basic approach.
  A \stylegan generator is trained on a dataset, \eg FFHQ~\cite{karras_style-based_2019}.
  Afterwards, we train the encoder part of an autoencoder for reconstruction and denoising tasks on a \emph{different} dataset, without updating the pre-trained \stylegan which is used as a decoder.
  }
  \label{fig:teaser}
  \vspace{-0.2cm}
\end{figure}

\noindent
\acfp{GAN} are the current state-of-the-art models in the area of unconditional and conditional image generation.
They are applied in various computer vision areas, \eg, image-to-image-translation~\cite{isola_image--image_2016,zhu_unpaired_2017,huang_multimodal_2018,shen_towards_2019}, image superresolution~\cite{liu_multi-level_2018,zhang_plug-and-play_2020,shaham_singan_2019}, or unconditional generation of various image types~\cite{goodfellow_generative_2014,miyato_spectral_2018,brock_large_2019,karras_progressive_2018}.
Over time, image quality, resolution, and realism of synthesized images was improved by a large margin~\cite{goodfellow_generative_2014,miyato_spectral_2018,karras_progressive_2018,karras_style-based_2019}.
The \stylegan architecture~\cite{karras_style-based_2019,karras_analyzing_2019} is one of the current state-of-the-art models for unconditional image generation.
The architecture of \stylegan with its projection into a semantically meaningful latent space $\mathcal{W}$ and the usage of noise inputs for stochastic variation not only allows to generate a diverse range of images but also enables meaningful image editing operations.
Using recent GAN inversion techniques, it is also possible to perform similarly meaningful edit operations on embeddings of real images in the latent space~\cite{abdal_image2stylegan_2019,abdal_image2stylegan++_2019,pidhorskyi_adversarial_2020}.
We briefly analyze the related work in \autoref{sec:related_work}.
However, to the best of our knowledge, previous work has not provided complete answers for the following open research questions in the domain of GAN inversion:
\begin{enumerate*}[label={(\arabic*)}]
   % \item To what degree does the latent code influence the reconstruction result and what is the influence of the stochastic noise?
   % \item Is it possible to use the stochastic noise for more than only providing stochastic variations of the generated image?
   \item To what degree does the latent code influence the reconstruction result?
   \item Can the stochastic noise provide more than stochastic variations of the generated image?
   \item Can a generator model (\ie \stylegan) trained in one domain be used to effectively reconstruct images for a different domain?
   \item Can an encoder model trained in one domain be effectively used reconstruct images for a different domain?
\end{enumerate*}

In this work, we provide answers to all of these questions.
We introduce a novel conditional \ac{GAN} that uses the \stylegan generator architecture and a ResNet-based encoder model for image reconstruction and present strategies to maximize the semantic meaning of the latent code in \autoref{sec:method}.
During the training of our model, we use an off-the-shelf, pre-trained \stylegan generator as a decoder and train an encoder, while leaving the pre-trained weights of the generator untouched (see \autoref{fig:teaser}).
We show in \autoref{sec:experiments} that our approach is not only able to faithfully reconstruct input images from the domain the generator was trained on, but also from other domains.
For example, we show that a \stylegan generator pre-trained on the FFHQ dataset~\cite{karras_style-based_2019} and an encoder trained to reconstruct images from the FFHQ dataset is able to faithfully reconstruct images from several different LSUN datasets~\cite{yu_lsun_2015} (churches, cats, or bedrooms).
Furthermore, we provide in-depth insights into how a pre-trained \stylegan generator is able to reconstruct images from other data distributions.
% Furthermore, we introduce a method for fast embedding of images into the latent space of \stylegan, while keeping the quality of the reconstruction result at a high level.
We conclude our work in \autoref{sec:conclusion}.
Overall, our contributions can be summarized as follows:
\begin{enumerate*}[label={(\arabic*)}]
   \item The first approach for faithful cross-domain image reconstruction, based on a fixed generator model, tested on a large variety of images from several domains.
   \item Novel insights about the capabilities of noise inputs in \stylegan.
   \item A training scheme that increases the semantic meaning of the latent codes.
   \item A fast method that allows embedding of up to 40 images per second on a single GPU (NVIDIA RTX~2080~TI), which is much faster than other recent \ac{GAN} inversion models, \eg by \cite{guan2020collaborative}, which can process approximately 1.4 images per second.
   \item A practical application of our model in the area of image denoising, where we match other state-of-the-art models. %  without the need for specific changes to our model
\end{enumerate*}
Our code and trained models are available online\footnote{\url{https://github.com/Bartzi/one-model-to-reconstruct-them-all}}.

\section{Related Work}
\label{sec:related_work}

\noindent
\acp{GAN} have first been proposed by Goodfellow~\etal~\cite{goodfellow_generative_2014} in 2014.
Since then, they have been improved through different measures, such as training at different scales~\cite{karras_progressive_2018}, adding novel weight normalization techniques~\cite{miyato_spectral_2018} or generating high-resolution images over a diverse set of classes~\cite{brock_large_2019}.

A recent work by Karras~\etal~\cite{karras_style-based_2019} proposes a novel architecture inspired by a work on neural style transfer~\cite{huang_arbitrary_2017}.
They train their \stylegan architecture on their collected FFHQ dataset containing images of human faces to generate high-quality, realistic images of human faces.
Moreover, Karras~\etal propose several improvements regarding architecture and normalization methods for \stylegan in a more recent work~\cite{karras_analyzing_2019}.
In the following, we describe the previous art related to the two tasks covered in our work, image reconstruction and image denoising.

\subsection{Image Reconstruction (Embedding, Inversion)}

\noindent
Generative models usually operate on a latent code and/or random noise as an input to generate new images~\cite{goodfellow_generative_2014,karras_style-based_2019,karras_analyzing_2019}.
Previous work on \ac{GAN} inversion attempts to understand and interpret the underlying mechanisms of \acp{GAN} by embedding existing images into a \ac{GAN} architecture.
These works can be roughly divided into two categories.

On the one hand, a given image can be embedded into the latent space of a trained \ac{GAN} on a per-image basis~\cite{abdal_image2stylegan_2019,abdal_image2stylegan++_2019,zhu2016generative,creswell2018inverting}.
% These approaches randomly select an initial code for each image and optimize it individually.
These methods achieve very faithful reconstructions, but require optimization or training a model for each image making image embeddings of large-scale datasets infeasible.

On the other hand, there are works similar to the idea of autoencoder architectures (\eg~\cite{Kingma2014}) which learn an encoder network for embedding an image.
This approach is used \eg~in~\cite{perarnau2016invertible,bau2019seeing,guan2020collaborative}.
It is computationally efficient, since the learned encoder can retrieve an encoding for a given image.
However, learning a code with semantic meaning proves to be a challenge as stated by Zhu~\etal~\cite{zhu2020domain}.
% Their recent work attempts to embed a code with semantic meaning~\cite{zhu2020domain} for a single domain.

\subsection{Image Denoising}

\noindent
A typical application area for image reconstruction is image denoising, where the task is to remove noise from a given noisy image, to restore the original image.
Here, we focus on image denoising techniques based on deep neural networks, for more detailed information about image denoising research, please refer to the following survey papers~\cite{fan_brief_2019,goyal_image_2020}.

Several neural-network-based image denoising systems have been proposed in the past~\cite{jain_natural_2009,chen_trainable_2017,xie_image_2012,zhang_learning_2017,zhang_plug-and-play_2020,zhang_ffdnet_2018,liu_multi-level_2018,zhang_beyond_2017}.
Some works have been trained for image denoising at fixed noise level~\cite{chen_trainable_2017}, while others are able to denoise noisy images with various noise levels.
Over time, the most common approach shifted from directly predicting the denoised image to predicting residual/noise images which are then subtracted from the input image, returning the denoised image~\cite{liu_multi-level_2018,zhang_beyond_2017}.
Based on this residual prediction strategy, further enhancements have been proposed.
Zhang \etal propose multiple extensions, \ie using multiple \acp{CNN} based denoising networks and model based optimization~\cite{zhang_learning_2017}, providing an extra noise level map as auxiliary input to the neural network~\cite{zhang_plug-and-play_2020,zhang_ffdnet_2018}, or creating specific network architectures for iamge denoising~\cite{zhang_plug-and-play_2020}.

% Talk a bit about image denoising, since this is an interesting application of our approach.

% \begin{itemize}
% 	\item has been actively researched in the past and survey papers are available~\cite{fan_brief_2019,goyal_image_2020}
% 	\item neural network based image denoising methods have been proposed
% 	\item either trained for specific one level of noise, or for a range of noise
% 	\item methods are specifically designed for image denoising
% 	\begin{itemize}
% 		\item usage of multiple \acp{CNN} based denoising networks and model based optimization for image denoising~\cite{zhang_learning_2017}
% 		\item specific mixtures of U-Net and ResNet building blocks~\cite{zhang_plug-and-play_2020}
% 		\item taking an extra noise level map as input~\cite{zhang_plug-and-play_2020,zhang_ffdnet_2018}
% 		\item predicting residual images that are applied to the original image~\cite{liu_multi-level_2018,zhang_beyond_2017}
% 	\end{itemize}
% \end{itemize}

\section{Method}
\label{sec:method}

\begin{figure*}[t]
    \centering
    \includegraphics[width=\textwidth]{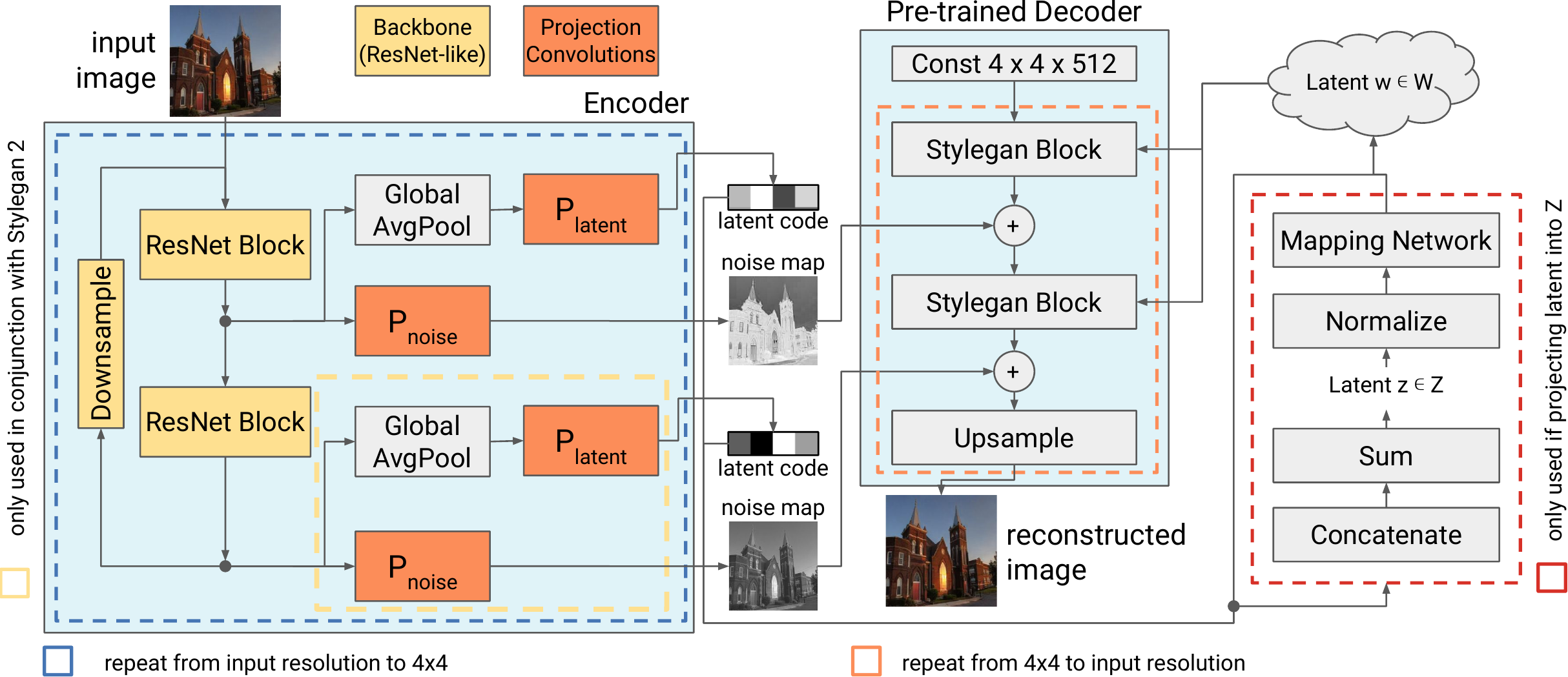}
    \caption{
    	An overview of the structure of our proposed autoencoder.
    	The autoencoder consists of an encoder and a pre-trained \stylegan 1 or 2 as decoder.
    	Our encoder consists of multiple ResNet blocks, each followed by convolutional layers that predict part of the latent code or a noise map.
    	Each output of the encoder is then used by the pre-trained \stylegan to reconstruct the input image.
    	As indicated in the figure, some parts of the model can be switched on or off, depending on the \stylegan version or latent projection strategy we use.
    }
    \label{fig:network_overview}
\end{figure*}

\noindent
In this section, we describe the method that we are using for the reconstruction of arbitrary input images (not limited to the training domain) by using a generative model that has only been trained for unconditional generation in a certain domain, \eg, face images (see \autoref{fig:teaser}).
In the following, we first introduce our overall architecture of the autoencoder that we use for image reconstruction (see \autoref{fig:network_overview}), split into the decoder (\autoref{sec:decoder}) and the encoder (\autoref{sec:encoder}).
During our experiments, we discovered how the encoder uses the noise maps instead of the latent code for image reconstruction (more details in \autoref{sec:results-noise}).
Therefore, we further developed two methods to reduce this behavior and instead maximize the semantic meaning of the latent code (see \autoref{sec:improve-latent-code}).
Finally, we describe the training details used in our experiments (see \autoref{sec:training-details}).

\subsection{Decoder Architecture}
\label{sec:decoder}

\noindent
For our experiments, we use generators based on \stylegan~\cite{karras_style-based_2019} and the improved version of \stylegan~\cite{karras_analyzing_2019} as our decoder network.
In the following we refer to the models based on the first version of \stylegan~\cite{karras_style-based_2019} as ``\stylegan 1'' and models based on the improved version of \stylegan~\cite{karras_analyzing_2019} as ``\stylegan 2''.

The \stylegan architecture currently sets the state-of-the-art in unconditional high-resolution image generation for a multitude of different natural image categories such as faces, buildings, and animals.
In order to generate high quality and high resolution images, \stylegan 1 and \stylegan 2 make use of a specialized generator architecture that is based on the idea of progressive growing for \acp{GAN}~\cite{karras_progressive_2018}.

All in all, the generator consists of three main components.
The first component is the mapping network that converts a latent vector $z \in \mathcal{Z}$ with $\mathcal{Z} \in \mathbb{R}^n$ into an intermediate latent space $w \in \mathcal{W}$ with $\mathcal{W} \in \mathbb{R}^n$.
The mapping network is implemented using a multilayer perceptron that typically consists of 8 layers with Leaky ReLU~\cite{maas_rectifier_2013} as activation function.
The resulting vector $w$ in that intermediate latent space is then transformed using learned affine transformations and used as an input to a synthesis network.

The second important component is stochastic noise, which is another input to the generator network.
Stochastic noise is added in the form of single channel images, where the value of each pixel is individually drawn from a normal distribution.
We denote these noise images as noise maps in the remainder of this paper.
In the original works~\cite{karras_style-based_2019,karras_analyzing_2019}, the authors add these noise inputs in order to generate stochastic detail.
However, in our experiments (see \autoref{sec:experiments}) we show that these noise maps can be used for even more than the generation of stochastic detail.

The third component is the synthesis network.
It consists of multiple blocks that each take three inputs.
First, they take a feature map that contains the current content information of the image that is to be generated.
Second, each block takes a transformed representation of the vector $w$ as an input to its style parts.
These style parts are implemented differently in the two versions of \stylegan.
In \stylegan 1 the \ac{AdaIN}~\cite{huang_arbitrary_2017} is used to guide the normalization of the feature map.
In \stylegan 2 \ac{AdaIN} is not used anymore, because \ac{AdaIN} leads to characteristic artifacts in images generated by \stylegan 1.
The artifacts are mitigated by the introduction of novel modulation and demodulation operations
The noise input is then added to the network after a normalization of the feature map.
The resulting feature map is then fed to the next block, which performs the same steps again and includes an upsample operation in every second block.

We note, that we use the proposed decoder architectures, \stylegan 1 and \stylegan 2, as is and without any changes.

\subsection{Encoder Architecture}
\label{sec:encoder}

\noindent
The second part of the autoencoder architecture is our encoder (see \autoref{fig:network_overview}).
The encoder is a fully convolutional network, that predicts latent vectors either in $\mathcal{Z}$ or in $\mathcal{W}$ and noise maps for each resolution of the generated images.
If we predict the latent vectors in $\mathcal{W}$, we follow the notion of \cite{abdal_image2stylegan_2019,abdal_image2stylegan++_2019} and name this space $\mathcal{W}^+$.
Our fully convolutional network follows a residual structure~\cite{he_deep_2016} and predicts latent vectors and all noise vectors following the U-Net architecture~\cite{ronneberger_u-net:_2015} (see the supplementary material for further details).
The predicted latent vectors and noise maps are then used as an input to a pre-trained \stylegan 1 or \stylegan 2 based generator, which is \emph{not} updated during the training process of the encoder.

\subsection{Maximizing the Semantic Meaning of the Latent Code}
\label{sec:improve-latent-code}

\noindent
We use a two-stage training scheme to increase the semantic meaning of the latent code.
The first stage disables learning (or usage) of the noise maps, forcing the model to only rely on the latent code for reconstruction.
The second stage then fine-tunes (or extends) the model to further improve the reconstruction results by training the layers responsible for predicting the noise maps.
We employ two different ways to achieve this goal.

\paragraph{Two Networks.} On the one hand, we use two independent encoder networks $L$ and $N$, where $L$ predicts the latent code and $N$ predicts the noise maps.
This allows us to train only $L$ during the first stage, while disabling $N$, thus forcing the model to use the latent code for the best result.
% During the second stage, this network predicting the latent code is not updated, instead the model learns to use the secondary network to predict.
In the second stage, we then use the pre-trained $L$ and only train $N$.

\paragraph{Learning Rate.} On the other hand, we use only one encoder model.
We adapt the learning rate or disable learning of the projection layers, $P_\mathrm{latent}$ and $P_\mathrm{noise}$, predicting the latent code and the noise maps, respectively (see \autoref{fig:network_overview}).
During the first stage, we only train $P_\mathrm{latent}$ (projecting into $\mathcal{Z}$ or $\mathcal{W}^+$) and do not train the $P_\mathrm{noise}$ layers.
In the second stage, we train all layers of our model, but we use different learning rates for different parts of the network.
$P_\mathrm{noise}$ layers are trained with the regular learning rate, whereas the learning rate for the backbone and $P_\mathrm{latent}$ is reduced, \eg, multiplied by $\frac{1}{100}$. \\
% Using this training strategy, our model is able to predict latent codes that have semantics, even though the generator has never been trained on such images.

The second method is more efficient, since it only requires one encoder network.
However, it is also susceptible to unlearn the usage of the latent code during stage two of the training if the learning rate is not tuned carefully.
We discuss the effect and results of both training strategies in \autoref{sec:results-semantic} and our supplementary material.

\subsection{Training Details and Loss Function}
\label{sec:training-details}

\noindent
% We train our proposed autoencoder architecture using stochastic gradient descent and on different datasets.
% Here, we want to stress again that we only train the weights of the encoder while keeping the weights of the pre-trained \stylegan fixed.
We use two different loss functions for the training of our models.
These loss functions are only used to update the weights of the \emph{encoder}, the weights of the decoder (a pre-trained \stylegan) are \emph{fixed}.
On the one hand, we use \ac{MSE} between the pixels of the generated image and the reconstructed image.
On the other hand, we utilize the \ac{LPIPS}~\cite{zhang_unreasonable_2018} metric for judging the reconstruction quality.
The resulting loss function is the following: $\mathcal{L}(x, y) = \mathcal{L}_\mathrm{mse}(x, y) + \mathcal{L}_\mathrm{lpips}(x, y).$
With $x, y \in \mathbb{R}^{[3, H, W]}$ being the input image and desired output image with three channels, height $H$ and width $W$, respectively.
$\mathcal{L}_\mathrm{mse}$ and $\mathcal{L}_\mathrm{lpips}$ denote \ac{MSE} and \ac{LPIPS} loss, respectively.

\section{Results and Discussion}
\label{sec:experiments}

% "internal" domain results
% Cross-domain results
% noise results

% two stage training results
% two-stem results
% one-stem results

% denoising results

\begin{table*}
    \begin{center}
    \begin{tabular}{l | rrr | rrr | rrr | rrr}
    
    \multirowcell{3}{\small Training Dataset,\\ \small \stylegan Version,\\ \small Projection Target} & \multicolumn{12}{c}{Dataset and Metric for Evaluation} \\
                                    & \multicolumn{3}{c|}{FFHQ}  & \multicolumn{3}{c|}{Church}      & \multicolumn{3}{c|}{Bedroom}       & \multicolumn{3}{c}{Cat}    \\
                                    & \fid   & \psnr   & \ssim   & \fid   & \psnr   & \ssim         & \fid   & \psnr   & \ssim           & \fid   & \psnr   & \ssim   \\
    \hline\hline
    \gc{}FFHQ, 1, $\mathcal{Z}$          &\gc9.85 &\gc25.03 &\gc0.91  &\gc7.17 &\gc20.37 &\gc0.88        &\gc3.74 &\gc23.32 &\gc0.91          &\gc4.17 &\gc22.14 &\gc0.88 \\
    \gc{}FFHQ, 1, $\mathcal{W}^+$        &\gc0.64 &\gc25.54 &\gc0.94  &\gc1.37 &\gc22.26 &\gc0.93        &\gc0.55 &\gc24.21 &\gc0.94          &\gc0.90 &\gc23.02 &\gc0.91 \\
    \gc{}FFHQ, 2, $\mathcal{Z}$          &\gc3.92 &\gc24.39 &\gc0.88  &\gc4.66 &\gc19.87 &\gc0.82        &\gc3.24 &\gc22.71 &\gc0.86          &\gc4.47 &\gc21.50 &\gc0.84 \\
    \gc{}FFHQ, 2, $\mathcal{W}^+$        &\gc0.75 &\gc29.11 &\gc0.95  &\gc1.23 &\gc24.48 &\gc0.94        &\gc0.57 &\gc28.13 &\gc0.96          &\gc0.96 &\gc26.01 &\gc0.93 \\
    \hline
    Church, 1, $\mathcal{Z}$        & 17.28  & 19.83   & 0.86    & 3.17   & 23.32   & 0.91          & 4.22   & 21.90   & 0.90            & 4.98   & 20.50   & 0.86 \\
    Church, 1, $\mathcal{W}^+$      &  3.30  & 20.99   & 0.90    & 0.26   & 26.18   & 0.95          & 1.25   & 23.65   & 0.94            & 1.37   & 22.15   & 0.90 \\
    Church, 2, $\mathcal{Z}$        & 12.24  & 20.92   & 0.83    & 3.17   & 23.08   & 0.86          & 9.76   & 22.04   & 0.85            & 7.33   & 20.76   & 0.82 \\
    Church, 2, $\mathcal{W}^+$      &  2.33  & 23.10   & 0.91    & 0.21   & 28.99   & 0.95          & 0.60   & 26.43   & 0.96            & 1.06   & 24.35   & 0.91 \\
    \hline
    Bedroom, 1, $\mathcal{Z}$       &  5.82  & 23.14   & 0.91    & 2.23   & 23.41   & 0.92          & 1.13   & 26.94   & 0.95            & 2.11   & 23.71   & 0.90 \\
    Bedroom, 1, $\mathcal{W}^+$     &  1.60  & 24.10   & 0.91    & 1.16   & 23.00   & 0.93          & 0.30   & 26.22   & 0.95            & 0.79   & 23.86   & 0.91 \\
    Bedroom, 2, $\mathcal{Z}$       &  7.17  & 22.70   & 0.87    & 6.48   & 20.49   & 0.84          & 2.96   & 24.29   & 0.88            & 5.10   & 21.97   & 0.85 \\
    Bedroom, 2, $\mathcal{W}^+$     &  1.59  & 26.44   & 0.93    & 1.06   & 26.97   & 0.94          & 0.36   & 31.01   & 0.97            & 0.84   & 26.78   & 0.93 \\
    \hline
    Cat, 1, $\mathcal{Z}$           & 39.65  & 24.60   & 0.92    & 6.68   & 24.25   & 0.93          & 4.01   & 26.71   & 0.94            & 2.96   & 25.03   & 0.92   \\
    Cat, 1, $\mathcal{W}^+$         &  1.12  & 24.94   & 0.93    & 0.88   & 24.29   & 0.94          & 0.28   & 26.40   & 0.96            & 0.57   & 24.86   & 0.93   \\
    Cat, 2, $\mathcal{Z}$           &  5.31  & 23.61   & 0.90    & 2.83   & 22.73   & 0.90          & 2.18   & 24.49   & 0.92            & 2.71   & 23.31   & 0.89   \\
    Cat, 2, $\mathcal{W}^+$         &  1.34  & 28.38   & 0.95    & 0.55   & 27.99   & 0.95          & 0.33   & 30.35   & 0.97            & 0.68   & 28.41   & 0.94   \\
    \end{tabular}
    \end{center}
    \caption{
        The results of our reconstruction experiments.
        We use a \stylegan model for decoding, which was pre-trained on the FFHQ dataset~\cite{karras_style-based_2019} and is not updated during the training of the encoder.
        Only the first highlighted row uses an encoder which is also trained on FFHQ, the other encoders are trained on different LSUN datasets \cite{yu_lsun_2015}.
        The first column shows the dataset, the version of \stylegan (1, 2), and the projection target ($\mathcal{Z},\mathcal{W}^+$).
        Each model is evaluated on different datasets and we report \ac{FID}, \ac{PSNR}, and \ac{SSIM}.
        }
    \label{tab:ffhq_reconstruction_results}
\end{table*}

\noindent
In this section, we show experimental results of our approach on different datasets and two different tasks, image reconstruction and image denoising.
% The idea behind the approach that we introduce in this paper is to investigate what a pre-trained \stylegan model is capable of.
% In this sense, we investigate whether a pre-trained \stylegan model is able to only generate images from the domain it has been trained on or not.
% In this section, we present a large range of experiments that show what is possible with a pre-trained \stylegan model.
First, we show that we are able to faithfully reconstruct images with our presented autoencoder architecture with a \stylegan decoder pretrained on the FFHQ dataset~\cite{karras_style-based_2019} and an encoder trained on the same dataset.
Second, we show the results for cross-domain reconstruction, where the encoder is trained on a different dataset, but the same pre-trained \stylegan decoder (trained on FFHQ) is used.
Third, we investigate how high-quality reconstruction is possible in this setting by examining the role of the noise maps in \stylegan.
% Here, we will also show that this observation holds true even for \stylegan models pre-trained on other datasets than the FFHQ dataset.
% Third, we use  well a pre-trained \stylegan model is able to perform semantic interpolations, even when not trained on a given class.
Afterwards, we present the results for our two-stage training method to increase the semantic meaning of the latent code.
Finally, we show the capabilities of our model when applied for the task of image denoising.

\subsection{Experimental Setup}

% \noindent
% We implement our model using PyTorch~\cite{paszke_pytorch_2019}.
% We use the FFHQ dataset~\cite{karras_style-based_2019} for high-quality images of human faces, and three LSUN datasets~\cite{yu_lsun_2015}, each of which contains only images of one type: churches, bedrooms, and cats.
% For all of our experiments, we train a model for \num{100000} iterations, using two GPUs with a batch size of \num{4} per GPU.
% We perform our experiments on a range of different GPUs with at least \SI{11}{\giga\byte} of GPU memory.
% We use the Adam~\cite{kingma_adam:_2015} optimizer with a cosine annealing learning rate schedule~\cite{DBLP:conf/iclr/LoshchilovH17} and an initial learning rate of \num{0.0001}.
% During the preprocessing, all input images are resized to $256 \times 256$ pixels, disregarding aspect ratios.
% We use the following three metrics on the given validation sets to evaluate our models:
% \begin{enumerate*}[label={(\arabic*)}]
%     \item The \ac{FID}~\cite{salimans_improved_2016} of reconstructed images with the original images (using a sample size of \num{50000} images).
%     Furthermore, we calculate
%     \item \ac{PSNR} and
%     \item \ac{SSIM}~\cite{zhou_wang_image_2004} between each input and its corresponding reconstructed image to measure the reconstruction quality.
% \end{enumerate*}

\noindent
We implement our model using PyTorch~\cite{paszke_pytorch_2019}.
We use the FFHQ dataset~\cite{karras_style-based_2019} for high-quality images of human faces, and three LSUN datasets~\cite{yu_lsun_2015}, each of which contains only images of one type: churches, bedrooms, and cats.
We use the following three metrics on the given validation sets to evaluate our models:
\begin{enumerate*}[label={(\arabic*)}]
    \item The \ac{FID}~\cite{salimans_improved_2016} of reconstructed images with the original images (using a sample size of \num{50000} images).
    Furthermore, we calculate
    \item \ac{PSNR} and
    \item \ac{SSIM}~\cite{zhou_wang_image_2004} between each input and its corresponding reconstructed image to measure the reconstruction quality.
\end{enumerate*}
Further details, such as, system details, number of iterations, optimizer, learning rate, and data preprocessing can be found in the supplementary material.

\subsection{FFHQ-based Image Reconstruction}
\label{sec:ffhq-image-reconstruction}

\noindent
In our first set of experiments, we determined how well our autoencoder architecture (introduced in \autoref{sec:method}) is able to reconstruct images of the FFHQ dataset~\cite{karras_style-based_2019}, when using a \stylegan model pre-trained on the FFHQ dataset.
In this line, we trained a range of different encoders, using both \stylegan 1 and \stylegan 2 decoders.
Furthermore, we examined the influence of different latent code projection strategies.
On the one hand, we project into $\mathcal{Z}$ and use the mapping network for creating a single vector in $\mathcal{W}$.
On the other hand, we project into $\mathcal{W}^+$, using multiple latent vectors.

The quantitative results (see the highlighted first row in \autoref{tab:ffhq_reconstruction_results}) show, that our autoencoder is able to perform reconstruction for different datasets with overall high quality.
Note that even though both encoder and decoder were trained on FFHQ only, they show a high reconstruction quality when evaluated on other datasets, \eg, containing churches, bedrooms, or cats.
The similarity metrics PSNR and SSIM also clearly show the advantages of using the projection strategy $\mathcal{W}^+$ over $\mathcal{Z}$.

\begin{figure}
    \centering
    \includegraphics{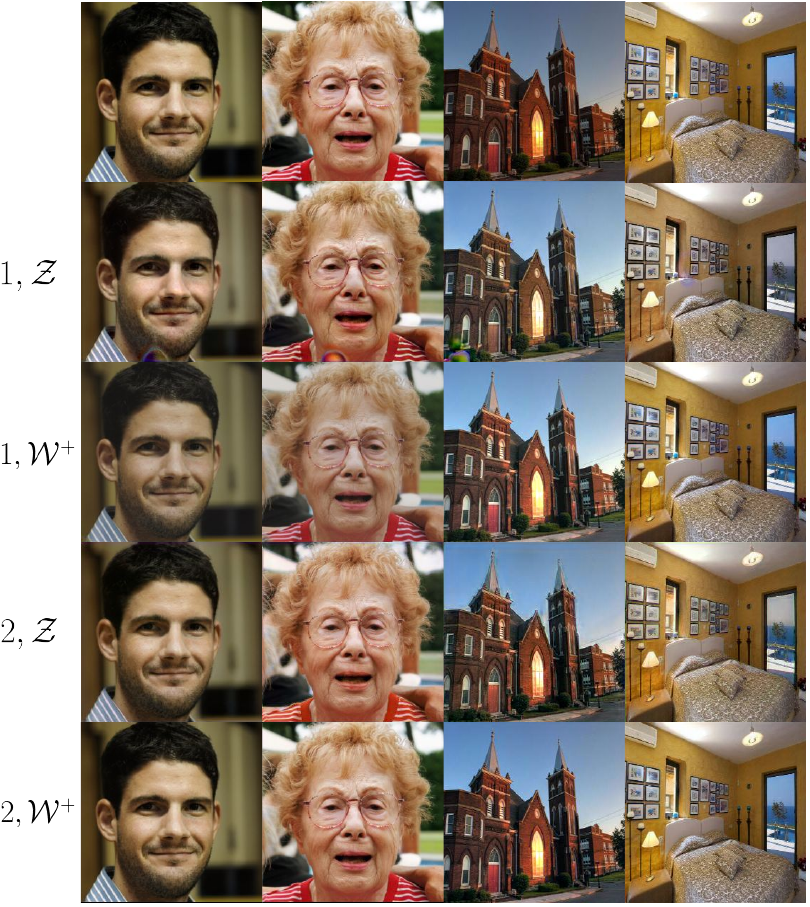}
    \caption{
    	Reconstruction results of our models trained with a \stylegan model pre-trained on the FFHQ dataset.
    	Images in the first row are real images, images in the following rows are reconstructions where the naming is as follows: \stylegan variant, latent projecting strategy.
    	Best viewed in color, more details visible when zoomed in.
    }
    \label{fig:ffhq_based_reconstruction_results}
\end{figure}

The qualitative results also show nearly no perceptual differences (see first two columns of \autoref{fig:ffhq_based_reconstruction_results}).
However, we can see that the models based on \stylegan 1 exhibit the ``bubble'' artifacts typical for images produced by \stylegan 1~\cite{karras_analyzing_2019}.
The absence of these artifacts in the reconstructed images of the \stylegan 2 based models is most likely the reason for the better quantitative results.

% The first row of \autoref{tab:ffhq_reconstruction_results} shows the results of our reconstruction experiments, where we train the encoder part of our autoencoder on the FFHQ dataset and evaluate the reconstruction capabilities of the resulting model on a range of different LSUN datasets~\cite{yu_lsun_2015}.

% From the results presented in \autoref{tab:ffhq_reconstruction_results} and \autoref{fig:ffhq_based_reconstruction_results}, we can see that our autoencoder is able to reconstruct images from different datasets with very high quality and nearly no perceptual differences.
% Such bubble artifacts are missing in all images produced by \stylegan 2 based models.
% This is most likely the reason, why the \stylegan 2 based models exhibit better results in our quantitative metrics.
% We note that the reconstruction results have a high quality even when the decoder is used for a different data distribution than it was trained for (\ie it was trained for human faces, but can reconstruct images of building and bedrooms).

\paragraph{Cross-Domain Image Reconstruction}
\label{sec:other-image-reconstruction}

\noindent
% In our second set of experiments, we determined how well our autoencoder architecture (introduced in \autoref{sec:method}) is able to reconstruct images of a dataset, when using a \stylegan model pre-trained on a \emph{different} dataset.
% As already shown in the last paragraph, our model is able to not only reconstruct images it has been trained on, but also images from other distributions.
Intrigued by our results on the FFHQ dataset, we trained a different set of autoencoder models that use the same pre-trained and fixed \stylegan generator, but use an LSUN dataset for training the encoder part of our autoencoder.
The quantitative results of these experiments are shown in rows 2-4 of \autoref{tab:ffhq_reconstruction_results}.
Further, we show the qualitative results of the encoders trained on churches and bedrooms in column 3 and 4 of \autoref{fig:ffhq_based_reconstruction_results}, respectively.
These results show that such cross-domain models are able to reconstruct images with comparable perceptual quality and scores.
We think that this is a very interesting result, since we are using a model trained for generating faces, but were able to use this model to generate virtually any image.
However, these finding are in line with the findings of Abdal \etal~\cite{abdal_image2stylegan_2019,abdal_image2stylegan++_2019}, who show that a latent code for many different images can be found, not only for images the model has been trained on.

\paragraph{The Significance of Noise for Image Reconstruction}
\label{sec:results-noise}

\begin{figure*}
\begin{center}
    \includegraphics[width=\linewidth]{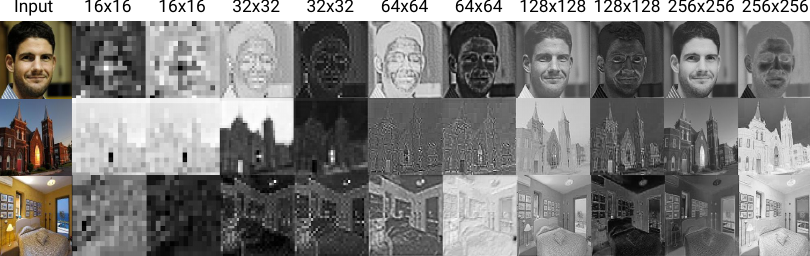}
    \caption{
    	Noise maps predicted by our model for the given input images on the left.
    	The noise maps are made visible by normalizing the values of each noise map independently.
    	It is clearly visible that all content information is saved in the noise maps.
    	The level of detail follows the progressive growing architecture of \stylegan and guides the generator by adding more and more content information to the already generated image.
    }
    \label{fig:predicted_noise_maps}
    \vspace{-0.3cm}
\end{center}
\end{figure*}

% After carefully investigating how our reconstruction models can produce such high quality results, we found that the model learns to use the noise inputs of \stylegan excessively for the reconstruction of input images and degrades the importance of the latent code at the same time.
% We could say that the role of latent codes and noise are swapped in our model.
% In \autoref{fig:predicted_noise_maps}, we show some examples of noise maps predicted by our encoder.
To understand how our reconstruction model can produce such high quality results, we examined the latent code and the noise maps predicted by our model.
First, we directly visualized the (normalized) noise maps predicted by our encoder (see \autoref{fig:predicted_noise_maps}).
It is clearly visible that the encoder learns to use the noise maps for retaining the content of the input image, especially in the noise maps of higher resolution.
This is an interesting observation, since the work of Karras et al. shows that the latent code provides semantic meaning \cite{karras_style-based_2019,karras_analyzing_2019} and the noise only provides semantically irrelevant details when used for a regular image generation task.
During reconstruction the roles seem effectively reversed, the latent code is barely used to store information, instead the noise maps capture the information up to a pixel level.

\begin{figure}
\begin{center}
    \includegraphics{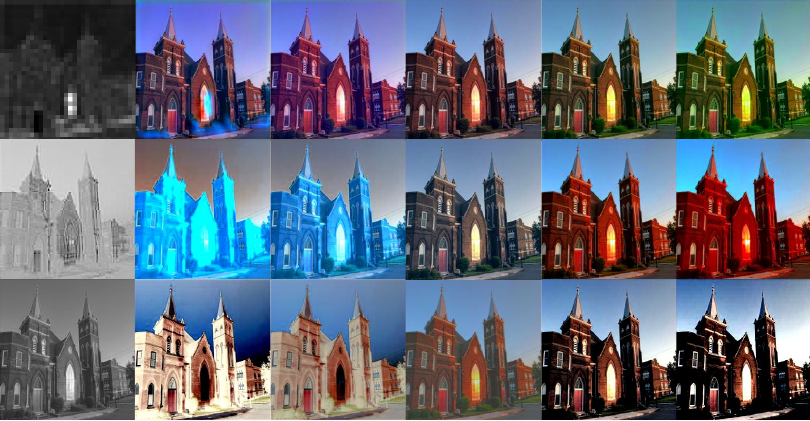}
    \caption{
    	Results of our experiments where we shifted the values of the predicted noise maps.
    	Each row shows the results when ``shifting'' each pixel of the noise map shown in the first column by multiplying the noise map with \num{-2}, \num{-0.75}, \num{0.5}, \num{1.75}, and \num{3} (from left to right).
    	It is clearly visible that the noise maps are not only used to encode the content of an image, but that noise can also be used to encode color and contrast of images.
        An extended version is in the supplementary material.
    }
    \label{fig:predicted_noise_maps_color_codes}
    \vspace{-0.3cm}
\end{center}
\end{figure}

To further analyze the meaning of the noise maps, we multiplied the value of each pixel in a noise map with a factor from the interval $\left[-2, 3 \right]$ and examined the reconstructed image.
The results (see \autoref{fig:predicted_noise_maps_color_codes}, an extended version is included in the supplementary material) show that the encoder uses the noise maps not only to capture the content of the image, but can also (at least to some degree) encode the colors of each pixel in these noise maps.

\subsection{Semantic Image Reconstruction}
\label{sec:results-semantic}

\begin{figure*}
% \captionsetup[subfigure]{justification=centering}
\begin{center}
\begin{subfigure}[t]{0.49\linewidth}
   \includegraphics[width=\linewidth]{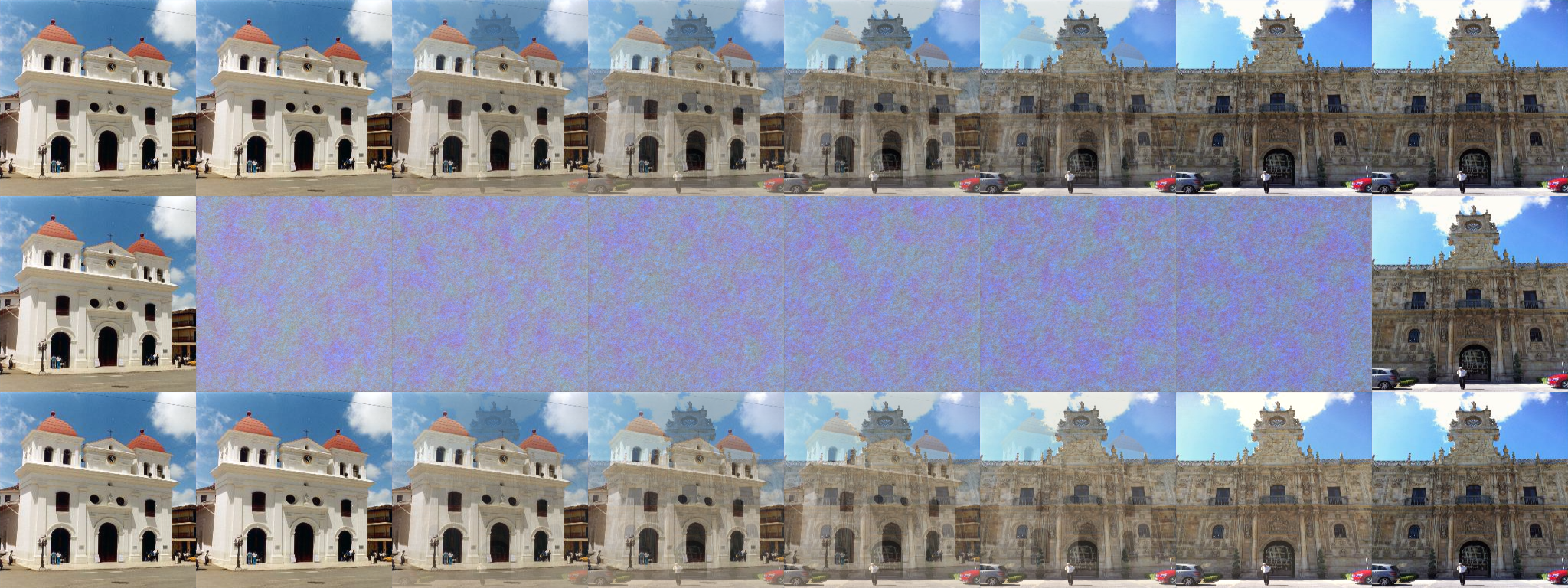}
   \caption{Regular training approach.}
   \label{fig:interpolation-w-plus}
\end{subfigure}
\hfill
\begin{subfigure}[t]{0.49\linewidth}
   \includegraphics[width=\linewidth]{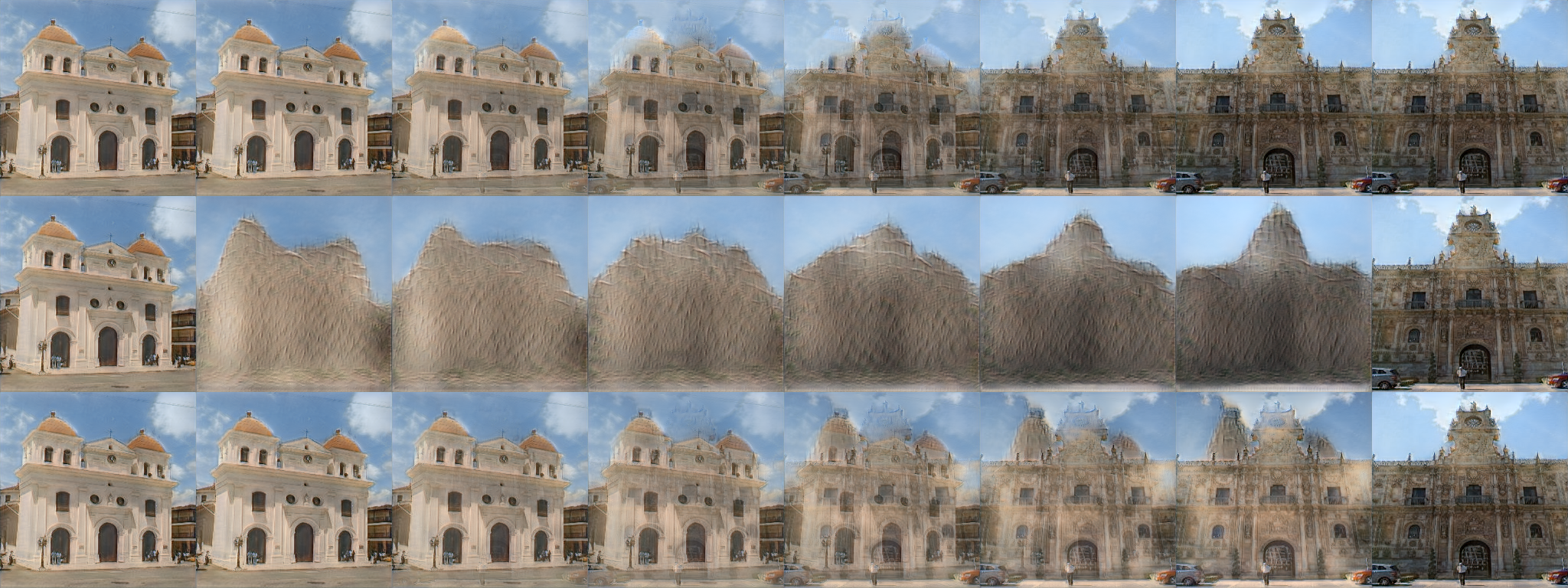}
   \caption{Two-stage training with two networks (see \autoref{sec:improve-latent-code}).}
   \label{fig:interpolation-two-stem}
\end{subfigure}
% \begin{subfigure}[t]{\linewidth}%{0.49\linewidth}
%    \centering
%    \includegraphics[width=\linewidth]{images/interpolations/lr_split}
%    \caption{Two-stage training with learning rate (see \autoref{sec:improve-latent-code}).}
%    \label{fig:interpolation-two-stem}
% \end{subfigure}
\end{center}
\vspace{-0.6cm}
    \caption{
    	Two images showcasing the behaviour of our models when interpolating latent code and noise maps between two reconstruction images based on two of our training strategies.
    	Both encoders embed into $\mathcal{W}^+$ of a \stylegan 2 decoder pre-trained on the FFHQ dataset.
    	% The right image shows a model that uses the same embedding approach, but was trained to be able to keep the semantics in the latent space.
    	The rows of each image show the following:
      The first row shows the interpolation between predicted latent code and noise at the same time.
    	The second row shows the interpolation of only the latent code, using a fixed random noise.
    	The third row shows the interpolation of only the noise maps, using the fixed latent code of the left image.
    }
    \label{fig:interpolation_results}
\end{figure*}

\noindent
% In the last paragraph, we examined the role of the noise maps for image reconstruction with our proposed autoencoder architecture.
% Based on this knowledge the question what role the latent code in this setting has, arises.
% Is it still possible to use the latent code and perform semantically meaningful interpolations between images?
Similarly, we examined the semantic meaningfulness of the latent code with sample interpolations between two images (see \autoref{fig:interpolation-w-plus}) based on a \stylegan 2 decoder trained on FFHQ.
This visualization shows, that a model simply trained for reconstruction is not able to perform semantic interpolation.
It is visually more similar to an alpha blending between two images.
Thus, it seems the influence of the latent code is degraded in such a way that it is only used to provide some basic colors for the resulting image and the resulting reconstruction completely depends on the predicted noise maps.

However, we also tested our improved two-stage training strategy (introduced in \autoref{sec:improve-latent-code}) to find a more meaningful (semantic) latent code and perform more meaningful semantic interpolations.
% Our new training strategy involves training the network in two steps and splitting one network into two networks.
% First, we train a network that only predicts the latent code ($\mathcal{Z}$ or $\mathcal{W}^+$), providing us with a model that is able to produce meaningful latent codes.
% In the second step, we train another network that only predicts the noise maps and use other the already trained model to supply the latent code.
% Using this training strategy, we are able to produce semantically meaningful latent codes, while retaining the reconstruction quality of our model.
% In the right sub-figure of \autoref{fig:interpolation_results} we show interpolation results of a model created with this training strategy.
% The results of the \emph{two network} strategy (see \autoref{fig:interpolation-two-stem}) show that a \stylegan generator trained on FFHQ is able to find semantically meaningful latent codes for images from other datasets that can be used for slight semantic interpolation, when trained with the proposed strategy.
The results of the \emph{two network} strategy (see \autoref{fig:interpolation-two-stem}) show that the semantic latent code captures the coarse structure of the content, but fine details are still added by the predicted noise maps.
We also observe that the interpolations seem to be more reasonable, but the visual quality of images reconstructed by using only the latent code is still visibly lower than the original images.
When incorporating the predicted noise maps on top of the semantic latent codes, the quality increases, but can not achieve the same level compared to our other experiments (the metrics and visualized interpolations for these models are included in the supplementary material).
The results of the \emph{learning rate} strategy can achieve similar results (see the supplementary material for details), however we found it can be tedious to find the optimal learning rates.
Without careful tuning of the actual learning rate in stage 2, the model unlearns the previously learned semantic meaning of the latent code.

\subsection{Reconstruction Speed}

\noindent
We reproduced the approach of Abdal \etal~\cite{abdal_image2stylegan_2019,abdal_image2stylegan++_2019} to measure its speed and found it needs approximately 7 minutes for a \emph{single} image on a RTX~2080~TI GPU.
The approach of Guan \etal~\cite{guan2020collaborative} is much faster, but still needs about 0.71 seconds per image on a Tesla V100 GPU, meaning they can process about 1.4 images per second.
For comparison, our model can process approximately 40 images per second on a RTX~2080~TI GPU, while still producing reconstructions with very high perceptual quality.

\subsection{Image Denoising}

\noindent
% Now that we understood, how our proposed model is able to reconstruct input image, we want to introduce our experimental results for an example application of our approach.
% A natural choice for the application of our the proposed model are image reconstruction tasks, as for instance image denoising, or image inpainting.
As a second task, we tested the capabilities of our approach on the exemplary application of image denoising.
% For this paper, we chose to examine the image denoising capabilities of our proposed model.
As already introduced in \autoref{sec:related_work}, several deep learning based image denoising techniques have been proposed in the past.
% Our proposed model has the advantage that it was not specifically designed as an image denoising network, unlike approaches from related work.
We note that our model was not specifically designed as an image denoising network, unlike approaches from other related work.
Compared to other works, our model harnesses the generative power of \stylegan to directly predict the denoised image without the need for a residual image that is subtracted from the noisy input image.

\begin{figure}
    \centering
    \includegraphics{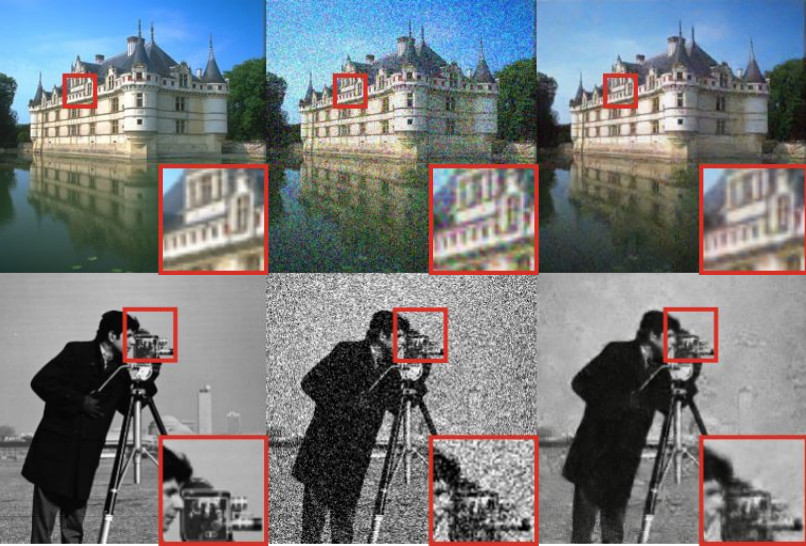}
    \caption{
    	Qualitative results of our model on the image denoising task.
    	We show two examples.
    	The example on the top is from the BSD68 dataset and the example on the bottom from the Set12 dataset.
    	For denoising, we used an autoencoder trained for denoising with an encoder embedding into the $\mathcal{W}^+$ space of a \stylegan 2 pre-trained on FFHQ.
    	For creating the noisy image, we used additive gaussian white noise with a standard deviation of \num{50}.
    }
    \label{fig:qualitative_image_denoising_results}
\end{figure}

\begin{table*}
    \begin{center}
    \begin{tabular}{l | ccc | ccc}
    Dataset                                 & \multicolumn{3}{c|}{Set12}                                & \multicolumn{3}{c}{BSD68}                                 \\
    \hline
    $\sigma$                                & 15                & 25                & 50                & 15                & 25                & 50                \\
    \hline\hline
    Liu~\etal~\cite{liu_multi-level_2018}   & 33.15/\bf{0.90}   & 30.79/\bf{0.87}   & 27.74/\bf{0.81}   & 31.86/0.90        & 29.41/0.84        & 26.53/0.74        \\
    Zhang~\etal~\cite{zhang_ffdnet_2018}    & 32.75/ -          & 30.43/ -          & 27.32/ -          & 31.63/ -          & 29.19/ -          & 26.29/ -          \\
    Zhang~\etal~\cite{zhang_plug-and-play_2020}& \bf{33.25}/ -  & \bf{30.94}/ -     & \bf{27.90}/ -     & \bf{31.91}/ -     & \bf{29.48}/ -     & \bf{26.59}/ -     \\
    \stylegan 1, $\mathcal{Z}$              & 24.67/0.82        & 24.13/0.76        & 22.27/0.60        & 24.59/0.86        & 24.48/0.83        & 23.71/0.73        \\
    \stylegan 1, $\mathcal{W}^+$            & 26.42/0.87        & 25.93/0.81        & 24.18/0.64        & 25.17/0.91        & 25.02/0.87        & 24.78/0.77        \\
    \stylegan 2, $\mathcal{Z}$              & 26.02/0.84        & 25.58/0.80        & 23.88/0.67        & 24.70/0.85        & 24.76/0.83        & 24.30/0.75        \\
    \stylegan 2, $\mathcal{W}^+$            & 27.46/0.88        & 27.08/0.85        & 24.94/0.71        & 27.57/\bf{0.92}   & 27.46/\bf{0.89}   & 26.22/\bf{0.81}
    \end{tabular}
    \end{center}
    \vspace{-0.2cm}
    \caption{
        Average \ac{PSNR}/\ac{SSIM} results of our model compared to state-of-the-art image denoising models.
        \textbf{Bold font} indicates the best performing result.
    }
    \label{tab:quantitative_results_image_denoising}
\end{table*}

For image denoising, we trained multiple models and compare them to the state-of-the-art in image denoising on the BSD68~\cite{roth_fields_2009} and the SET 12~\cite{martin_database_2001} benchmark datasets.
We trained models based on \stylegan 1 and \stylegan 2 (both pre-trained on the FFHQ dataset), with latent code embedding into $\mathcal{Z}$ and $\mathcal{W}^+$, and use the ImageNet dataset~\cite{deng_imagenet:_2009} for training the encoder.
We report the average \ac{PSNR} and \ac{SSIM} of our model on different noise levels $\sigma=15,25,\text{ and }50$ on the benchmark datasets in \autoref{tab:quantitative_results_image_denoising}.
The qualitative results can be seen in \autoref{fig:qualitative_image_denoising_results}.
Our results show that our model is able to perform image denoising at a level close to the state-of-the-art, even though our network has not been designed with the application of image denoising in mind.
Some ideas mentioned in related word (\eg network design decisions, similar to~\cite{liu_multi-level_2018} or \cite{zhang_plug-and-play_2020}) could be used to boost the performance of our model, but such improvements are out of the scope of this work, since here we only want to show that such an application is possible.

The results are nonetheless quite interesting considering the generator of the network has never been trained for image reconstruction and also never for the creation of images apart from faces of the FFHQ dataset.
Another observation based on the qualitative results of the denoised images is that our model not only removes noise from an image, but also performs a slight shift in colors similar to a color correction operation.
Since changing the colors might be an unwanted result when removing noise from input images, the model might perform even better in the quantitative analysis, without this kind of color correction.
This behavior is an interesting property of our overall model and opens up further application possibilities of our proposed network in future work.

\section{Conclusion}
\label{sec:conclusion}

\noindent
In this work we showed a novel approach, which can achieve high-quality image reconstruction results.
In contrast to previous work, our model is highly efficient and can be applied for cross-domain image reconstruction.
We showed that a pre-trained \stylegan generator can be used to reconstruct images from virtually any dataset.
We provided an in-depth analysis of the causes for this behavior and found that the stochastic noise inputs, which are only meant to produce stochastic variations, can be used to capture small details, and manipulate colors of images generated by \stylegan without using the latent code.
Since we can circumvent using the latent code, the latent code loses its semantic expressiveness.
However, we proposed a two-stage training strategy for two slightly different architectures, which increases the semantic expressiveness of the computed latent code without decreasing the reconstruction quality by a large margin.
% However, we showed that we can still find semantically meaningful latent codes and produce high quality reconstructions, when making small changes to our architecture.

In future work, we would like to investigate further application possibilities of our proposed autoencoder architecture, \eg applications in the area of image-to-image translation.

\paragraph{Acknowledgement}

We wish to thank the Wildenstein Plattner Institute for supplying us with data and a research objective that ultimately led to the results published in this paper.

{\small
\bibliographystyle{ieee_fullname}
\bibliography{egbib}
}

\appendix
\section{Further Implementation Details}

\noindent
We implemented our model in PyTorch, as already stated in the main paper.
We based our implementation of the \stylegan 1 and \stylegan 2 models on freely on Github available re-implementations of \stylegan 1\footnote{\url{https://github.com/rosinality/style-based-gan-pytorch}} and \stylegan 2\footnote{\url{https://github.com/rosinality/stylegan2-pytorch}}.
Our code is also available on Github\footnote{\url{https://github.com/Bartzi/one-model-to-reconstruct-them-all}} and our training logs can be viewed online\footnote{\url{https://wandb.ai/hpi/One Model to Generate them All}}.

We perform our experiments on a range of different GPUs with at least \SI{11}{\giga\byte} of GPU memory.
For all of our experiments, we train a model for \num{100000} iterations, using two GPUs with a batch size of \num{4} per GPU.
We use the Adam~\cite{kingma_adam:_2015} optimizer with a cosine annealing learning rate schedule~\cite{DBLP:conf/iclr/LoshchilovH17} and an initial learning rate of \num{0.0001}.
During the preprocessing, all input images are resized to $256 \times 256$ pixels, disregarding aspect ratios.

\paragraph{Network Details}

Our encoder is based on the ResNet architecture, but does not follow the layout of other well-known ResNet feature extractors, \eg ResNet-18, or ResNet-152.
Instead, the number of convolutional layers in our feature extractors depends on the resolution of the input image.
The number of necessary ResNet blocks can be calculated using the following formula:
\begin{equation}
    \text{number of blocks} = 2 + 2 \cdot (\mathrm{log}_2(\text{insize}) - \mathrm{log}_2(\text{outsize})).
\end{equation}
We first start with two "start blocks", then we use two ResNet blocks for each resolution from input size to output size of the encoder.
The output size of the encoder is typically set to \num{4}, wheras we use \num{256} as our input size.
When using \num{256} and \num{4} as an input and output size, respectively, we get the following number of ResNet blocks:
\begin{align}
    \text{number of blocks} & = 2 + 2 \cdot (\mathrm{log}_2(256) - \mathrm{log}_2(4)) \\
                            & = 2 + 2 \cdot (8 - 2) \\
                            & = 2 + 2 \cdot 6 \\
                            & = 14.
\end{align}

Following each ResNet block, the network splits into three branches.
If we use \stylegan 1 as decoder, we only split after the first ResNet block of each resolution.
The first split is followed by a global average pooling and a $1\times1$ convolution that predicts parts of the latent code.
The second split is followed by a $1\times1$ convolution that is used to predict the noise map for the current feature map resolution.
The third branch goes to the next ResNet block.
Please see Figure 2 of the main paper for a structural overview.

\section{Experimental Results on Further Pre-Trained \stylegan Models}

\begin{figure*}
    \centering
    \begin{subfigure}{0.83\textwidth}
	    \includegraphics[width=1\linewidth]{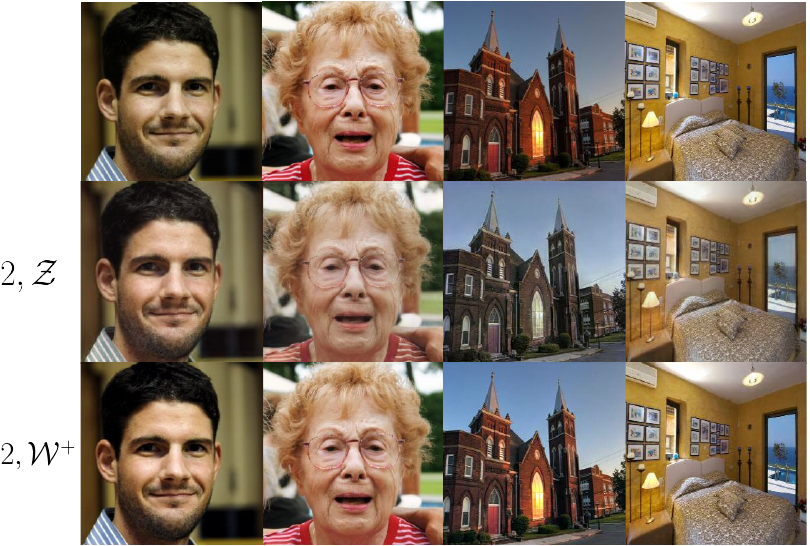}
	    \caption{
	    	Results for models based on \stylegan models pre-trained on the LSUN Cat dataset.
	    }
	    \label{fig:sup_cat_based_reconstruction_results}
	\end{subfigure}
	\begin{subfigure}{0.83\textwidth}
		\centering
	    \includegraphics[width=1\linewidth]{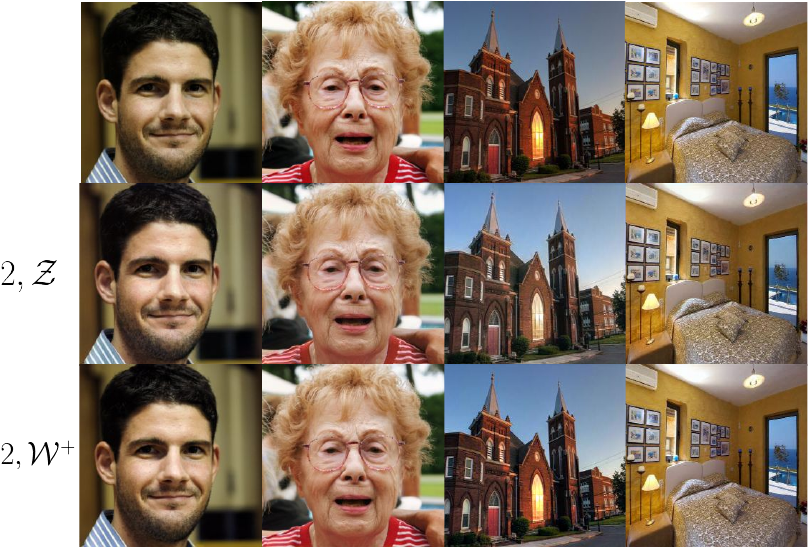}
	    \caption{
	    	Results for models based on \stylegan models pre-trained on the LSUN Church dataset.
	    }
	    \label{fig:sup_church_based_reconstruction_results}
	\end{subfigure}
	\caption{
	 	Results of further Reconstruction experiments with \stylegan models pre-trained on other datasets than FFHQ.
	 	Images in the first row are real images, images in the following rows are reconstructions where the naming is as follows: \stylegan variant, latent projecting strategy.
	    Best viewed in color.
	}
\end{figure*}

\noindent
In order to show that reconstruction with \stylegan does not only work with a generator pre-trained on the FFHQ dataset, we also performed experiments with generators pre-trained on the datasets LSUN church and LSUN cat.
The results of these experiments show that our approach is also applicable to other pre-trained generators and reaches a very similar performance on a range of different datasets.
Please note, that here we only experimented with models based on \stylegan 2, since we were not able to get pre-trained models for the Pytorch implementation of \stylegan 1 that we base our work on, but we are absolutely certain that we can reach the same results with models based on \stylegan 1.
We show the quantitative results of our experiments in \autoref{tab:sup_cat_based_eval_results} and \autoref{tab:sup_church_based_eval_results}.
Furthermore, we also show more image reconstruction results in \autoref{fig:sup_cat_based_reconstruction_results} and \autoref{fig:sup_church_based_reconstruction_results}.

\section*{Further Visualizations Explaining the Role of Noise}

\begin{figure*}[p]
    \centering
    \includegraphics{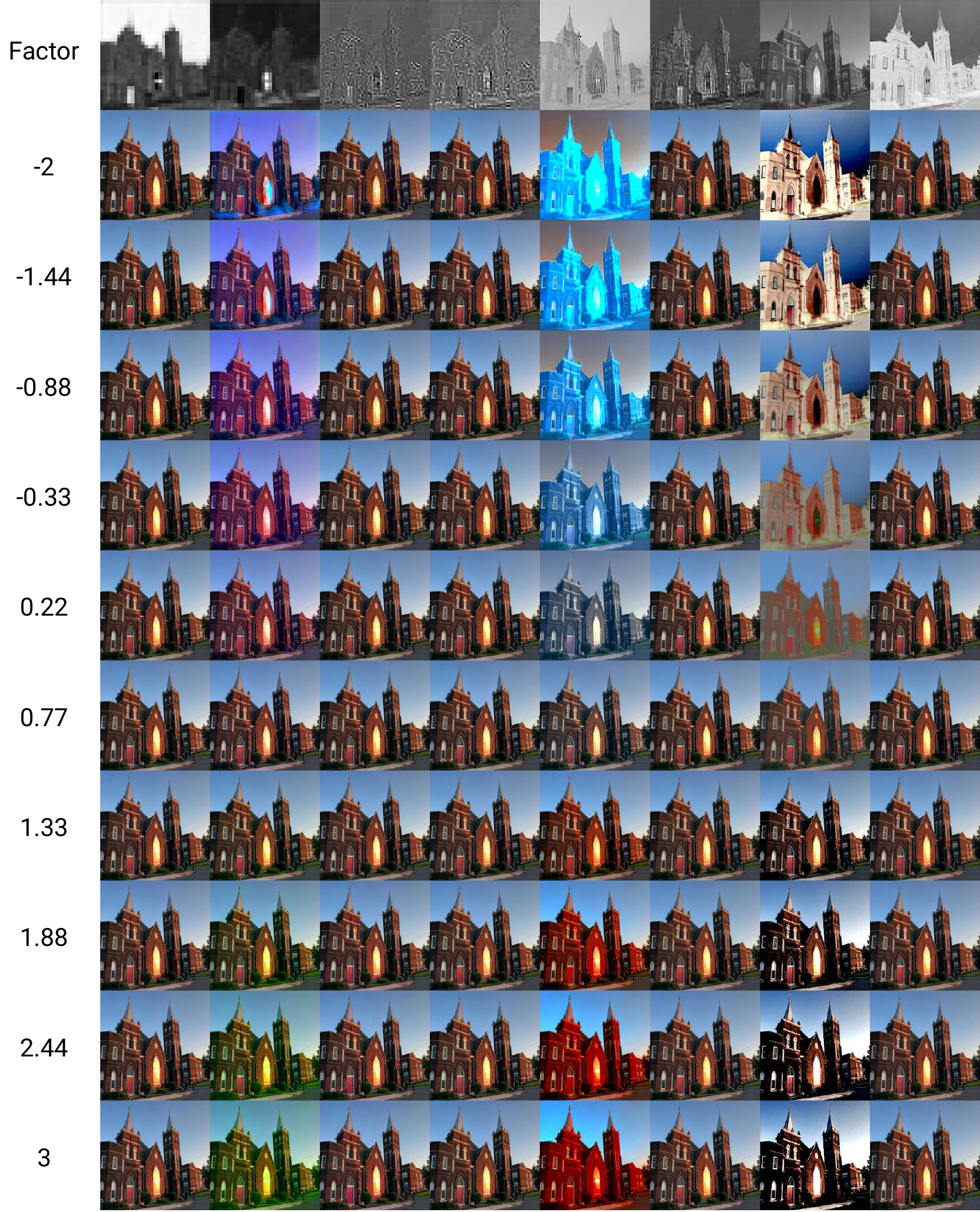}
    \caption{
    	Longer version of noise shifting experiments with a \stylegan 2 generator pre-trained on the FFHQ dataset.
    	Each column shows the results when ``shifting'' each pixel of the corresponding noise map shown in the first row of the column by multiplying the noise map with the factors indicated at the left side.
    	It is clearly visible that the noise maps are not only used to encode the content of an image, but that noise can also be used to encode color and contrast of images.
    	However, not all noise maps are necessary for this color coding, as the result does not change for some noise maps, regardless of the factor used.
    }
    \label{fig:sup_predicted_noise_maps_color_codes_ffhq_base}
\end{figure*}

\begin{figure*}[p]
    \centering
    \includegraphics{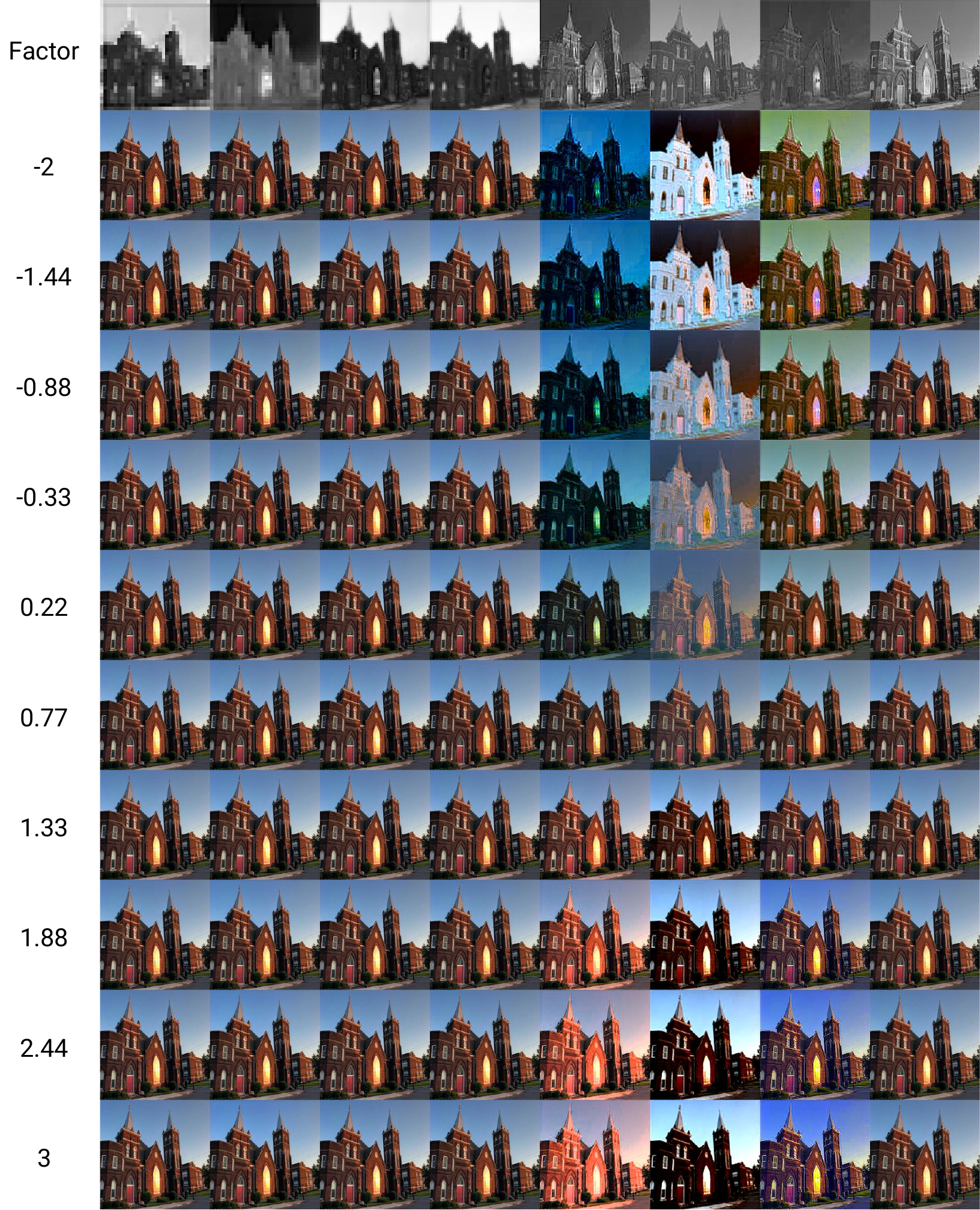}
    \caption{
    	Further results of noise shifting experiments.
    	Here we show the results with a \stylegan 2 generator pre-trained on the LSUN church dataset.
    	The semantics are the same as in \autoref{fig:sup_predicted_noise_maps_color_codes_ffhq_base}.
    	Here it is also clearly visible that the noise maps are not only used to encode the content of an image, but that noise is also used to encode color and contrast of images, but for this model different noise maps are used and the color model encoded is also different.
    }
    \label{fig:sup_predicted_noise_maps_color_codes_church_base}
\end{figure*}

\begin{figure*}[p]
    \centering
    \includegraphics{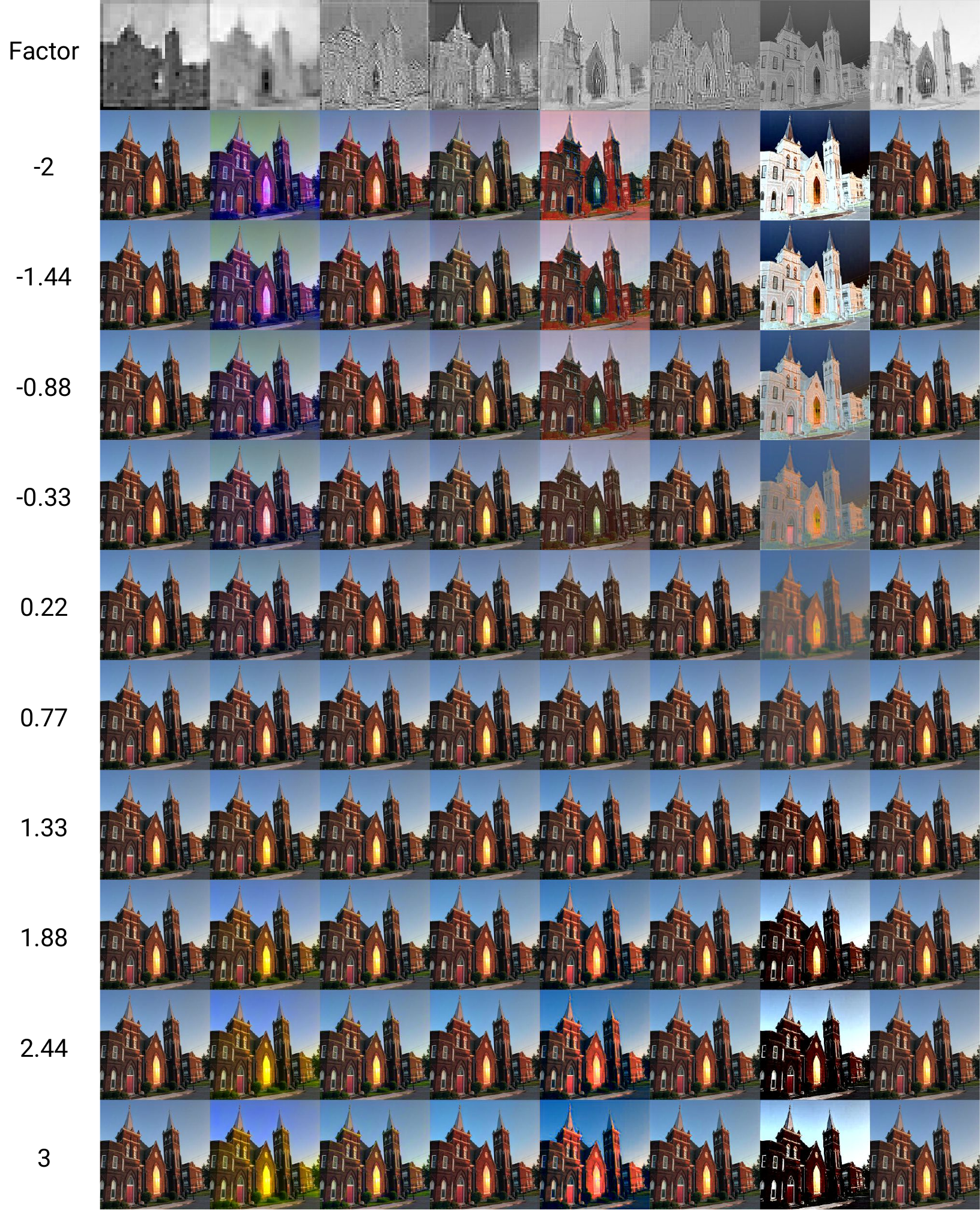}
    \caption{
    	Further results of noise shifting experiments.
    	Here we show the results with a \stylegan 2 generator pre-trained on the LSUN cat dataset.
    	The semantics are the same as in \autoref{fig:sup_predicted_noise_maps_color_codes_ffhq_base} and \autoref{fig:sup_predicted_noise_maps_color_codes_church_base}.
    	Here it is also clearly visible that the noise maps are not only used to encode the content of an image, but that noise is also used to encode color and contrast of images, but for this model different noise maps are used and the color model encoded is again different.
    }
    \label{fig:sup_predicted_noise_maps_color_codes_cat_base}
\end{figure*}

\noindent
In Figures \ref{fig:sup_predicted_noise_maps_color_codes_ffhq_base}, \ref{fig:sup_predicted_noise_maps_color_codes_church_base}, and \ref{fig:sup_predicted_noise_maps_color_codes_cat_base} we show more detailed results of our noise shifting experiments.
These experiments show that noise is used in several ways:
\begin{enumerate*}[label={(\arabic*)}]
    \item Noise can be used to control the content of the generated image.
    \item Noise can be used to control the colors of individual pixels of the generated image.
    \item A \stylegan model trained on one dataset can make use of noise in a different way than a \stylegan model trained on a different dataset, \ie the color coding depends on the \stylegan model and used dataset for training the \stylegan model.
\end{enumerate*}

\section*{Further Visualizations Showing Interpolation Results}

\begin{figure*}[p]
    \centering
    \begin{subfigure}{\textwidth}
        \includegraphics[width=1\linewidth]{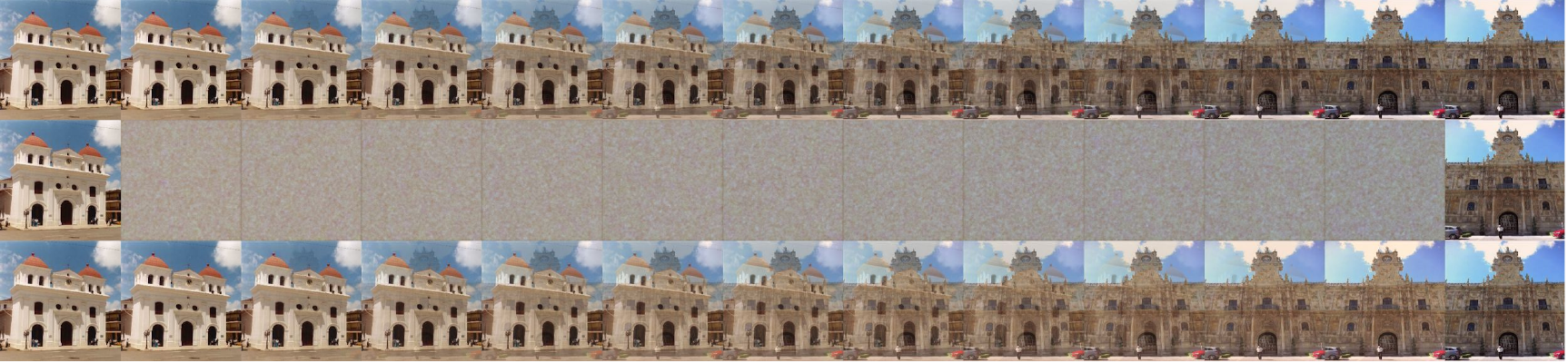}
        \caption{
            Encoder trained on FFHQ, decoder pre-trained on FFHQ, with \stylegan 2, and projecting into $\mathcal{W}^+$.
        }
        \label{fig:sup_interpolation_ffhq_stylegan_2_w_plus}
    \end{subfigure}
    \begin{subfigure}{\textwidth}
        \includegraphics[width=1\linewidth]{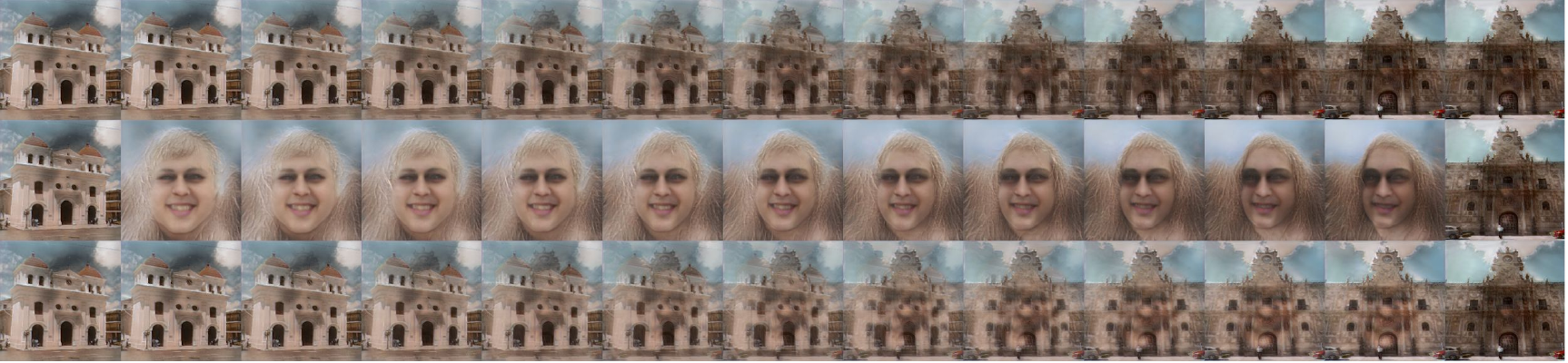}
        \caption{
            Encoder trained on FFHQ, decoder pre-trained on FFHQ, with \stylegan 2, projecting into $\mathcal{W}^+$, and our two network strategy.
        }
        \label{fig:sup_interpolation_ffhq_stylegan_2_w_plus_two_stem}
    \end{subfigure}
    \begin{subfigure}{\textwidth}
        \includegraphics[width=1\linewidth]{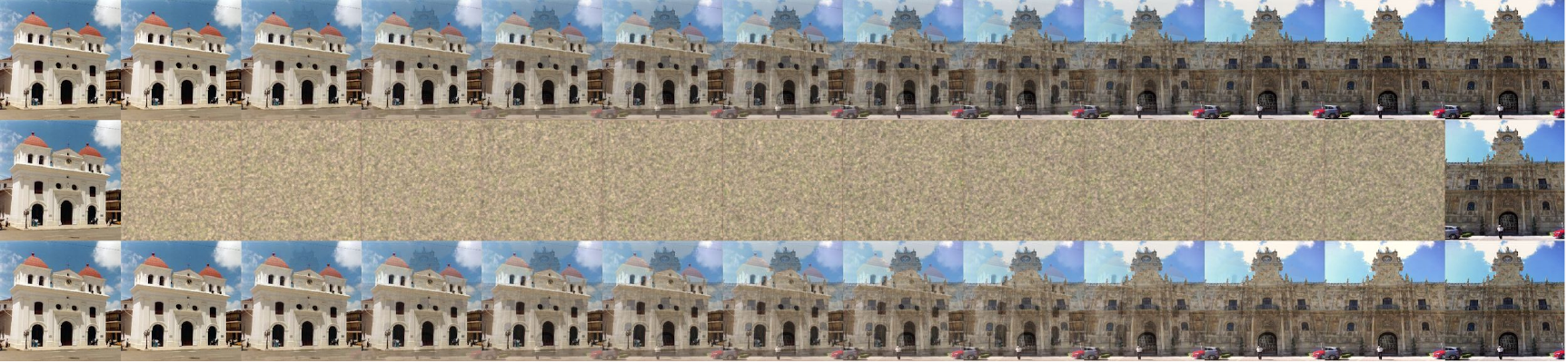}
        \caption{
            Encoder trained on LSUN cat, decoder pre-trained on FFHQ, with \stylegan 2, and projecting into $\mathcal{W}^+$.
        }
        \label{fig:sup_interpolation_ffhqlsun_cat_stylegan_2_w_plus}
    \end{subfigure}
    \begin{subfigure}{\textwidth}
        \includegraphics[width=1\linewidth]{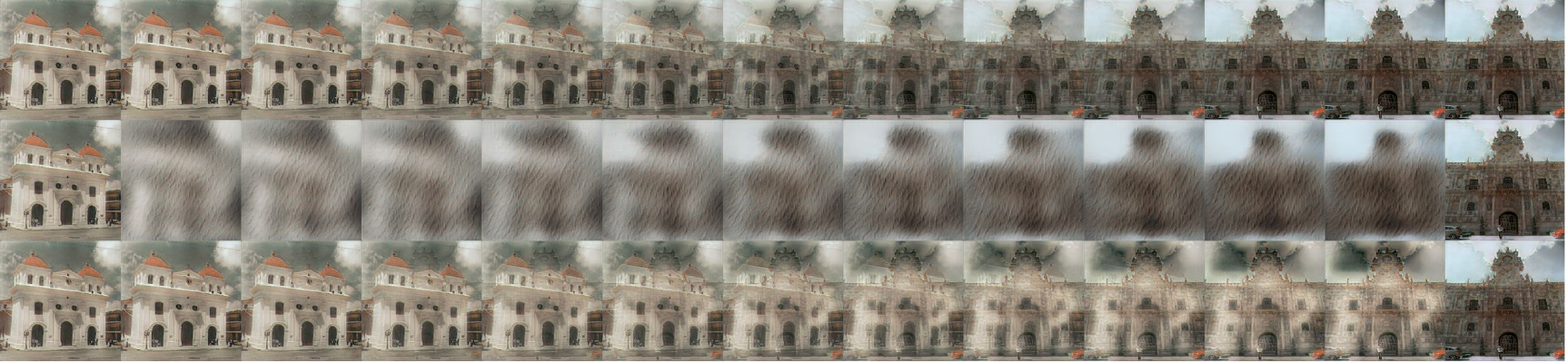}
        \caption{
            Encoder trained on LSUN cat, decoder pre-trained on FFHQ, with \stylegan 2, projecting into $\mathcal{W}^+$, and our two network strategy.
        }
        \label{fig:sup_interpolation_lsun_cat_stylegan_2_w_plus_two_stem}
    \end{subfigure}
    \caption{
        Several images showcasing the behaviour of our models when interpolating latent code and noise maps between two input images.
        We can observe that models trained without the two network strategy do not make any "intelligent" use of the latent code.
        The latent code is only used to encode some colors, as can be seen in the second-to-right image in the bottom-most row in \subref{fig:sup_interpolation_ffhq_stylegan_2_w_plus} and \subref{fig:sup_interpolation_ffhqlsun_cat_stylegan_2_w_plus}, since these images should show the target image, which is the right-most image in each row.
        For the other models, we can observe that our two network strategy can only be used to faithfully reconstruct images of the same data distribution.
        A semantically meaningful interpolation can be observed in \subref{fig:sup_interpolation_ffhq_stylegan_2_w_plus_two_stem}, but since the base model was trained on the FFHQ dataset, the latent code can only directly be used to generate faces.
        The results in \subref{fig:sup_interpolation_lsun_cat_stylegan_2_w_plus_two_stem} are similar to the results in the main paper, but here we can also see that noise is only rendered on top of the image generated by the latent code.
    }
    \label{fig:sup_interpolation_results}
\end{figure*}

\noindent
In \autoref{fig:sup_interpolation_results} we show further interpolation results.
Again, we can observe that models trained with the two network approach provide more meaningful interpolations.
We can also make the following observations:
\begin{enumerate*}[label={(\arabic*)}]
    \item If using a model trained on another domain than the input image, \eg a model trained to reconstruct FFHQ using the two network strategy and using images from the LSUN church dataset (see \autoref{fig:sup_interpolation_ffhq_stylegan_2_w_plus_two_stem}), we can see that the encoder explicitly learns to embed the latent code into the regions where faces can be generated, regardless of the input image.
    We can hence conclude that the encoder does not learn anything about the content of the image when embedding into the latent code.
    \item in \autoref{fig:sup_interpolation_ffhq_stylegan_2_w_plus_two_stem} and \autoref{fig:sup_interpolation_lsun_cat_stylegan_2_w_plus_two_stem}, we can observe that the predicted noise maps are ``rendered'' on top of the results of the latent code.
    This might be because the two networks in our two network architecture are independent of each other.
    In the future it might be worthwile to achieve a closer coupling of noise and latent code in the two network approach.
    \item Overall, we can observe that, when not training with the two network strategy, the latent code is only used to add color variations into the resulting images.
    This can, for instance, be observed in \autoref{fig:sup_interpolation_ffhq_stylegan_2_w_plus} in the second-to-right image in the bottom-most row.
    Here, we can see that the latent code also influences the colors of the resulting image, since the image on the right is the reconstruction of the second image without any interpolation between the two input images.
\end{enumerate*}

\begin{figure*}[p]
    \centering
    \begin{subfigure}{\textwidth}
        \includegraphics[width=1\linewidth]{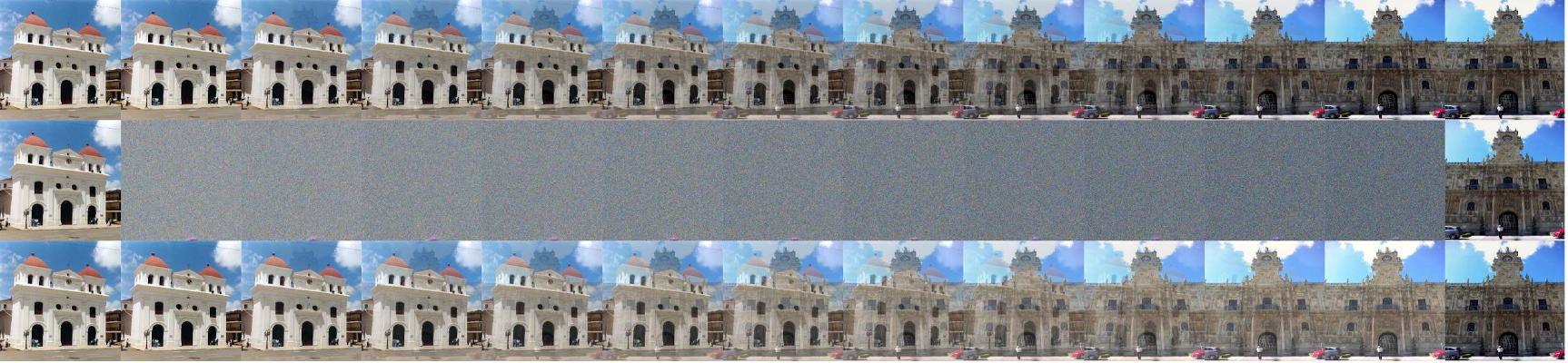}
        \caption{
            Encoder trained on LSUN church, decoder pre-trained on FFHQ, with \stylegan 1, and projecting into $\mathcal{W}^+$.
        }
    \label{fig:sup_interpolation_church_stylegan_1_w_plus}
    \end{subfigure}
    \begin{subfigure}{\textwidth}
        \includegraphics[width=1\linewidth]{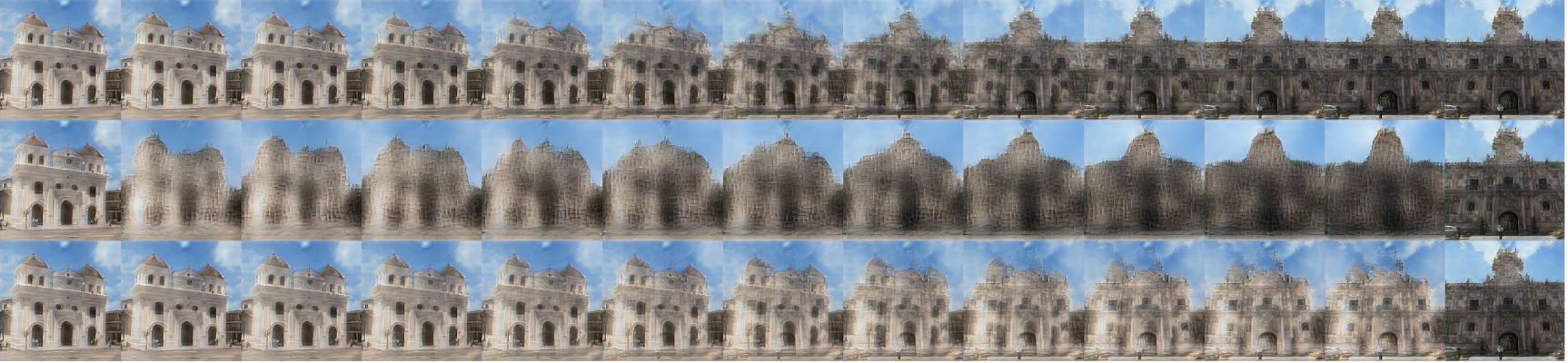}
        \caption{
            Encoder trained on LSUN church, decoder pre-trained on FFHQ, with \stylegan 1, projecting into $\mathcal{W}^+$, and our two network strategy.
        }
        \label{fig:sup_interpolation_church_stylegan_1_w_plus_two_stem}
    \end{subfigure}
    \begin{subfigure}{\textwidth}
        \includegraphics[width=1\linewidth]{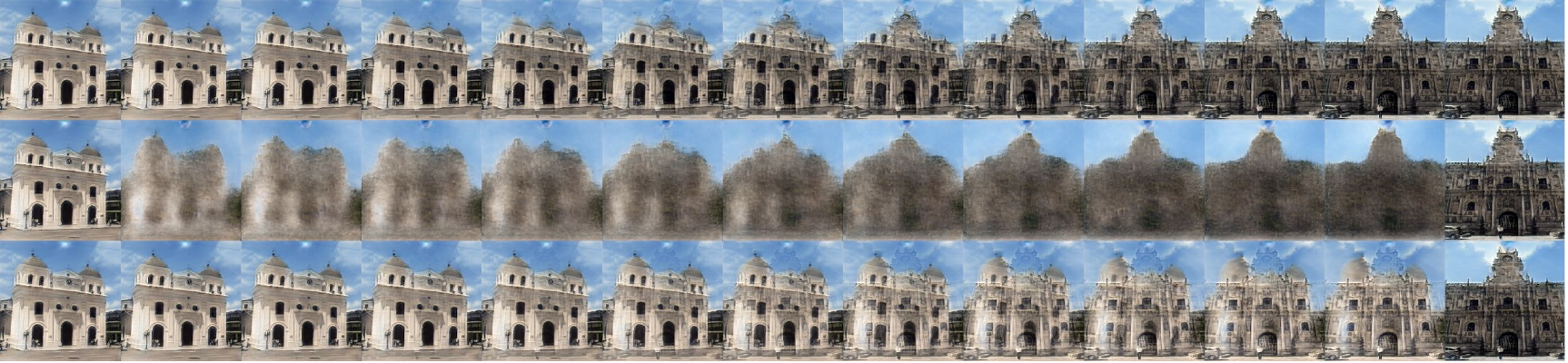}
        \caption{
            Encoder trained on LSUN church, decoder pre-trained on FFHQ, with \stylegan 1, projecting into $\mathcal{W}^+$, and using the learning rate strategy.
        }
        \label{fig:sup_interpolation_church_stylegan_1_w_plus_lr_strategy}
    \end{subfigure}
    \caption{
        Further images showcasing the behavior of our models when interpolating latent code and noise maps between two input images.
        Here, we used models trained on \stylegan 1 that project into $\mathcal{W}^+$.
        We compare plain projection into $\mathcal{W}^+$, our two network strategy and our learning rate strategy to maximise the semantic meaning of the predicted latent code.
        We can observe that the results of the two network and the learning rate strategy (see \subref{fig:sup_interpolation_church_stylegan_1_w_plus_two_stem} and \subref{fig:sup_interpolation_church_stylegan_1_w_plus}) are of similar visual quality.
        However, the reconstructions in \subref{fig:sup_interpolation_church_stylegan_1_w_plus} have a slightly better visual quality and also the interpolations seem to be more reasonable.
        We conclude that the learning rate strategy is superior to the two network strategy, but it is more difficult to find the correct learning rate ratio, as already discussed in the main paper.
    }
    \label{fig:sup_interpolation_results_two_stem_vs_lr_split}
\end{figure*}

Since, we also want to compare the two approaches for maximising the semantic meaning of the latent code, we provide a comparison of interpolations for the approaches based on the two network and the learning rate strategy in \autoref{fig:sup_interpolation_results_two_stem_vs_lr_split}.
We can observe that the learning rate split strategy produces better qualitative results.
However, it is very difficult to achieve these results for each \stylegan version, since the learning rate has to be tuned specifically.

\section{Quantitative Reconstruction Results of our Two Network Models}

\noindent
In \autoref{tab:sup_04_two_stem_quantitative} we show quantitative results for reconstruction on multiple datasets when using a model trained with the two network strategy.
The results show that a model trained with the two network strategy is still able to provide meaningful reconstructions.
However, the results also show that the resulting reconstructions are of worse quality than the reconstructions that nearly completely rely on the noise input.
Furthermore, we can see that the resulting models are not well suited for cross domain reconstruction.
We argue that this is because the latent code is specialized to a specific dataset/data distribution and is not able to handle inputs that are different, since the encoder has learned to project the input image into the regions of the latent space for the specific type of data it has been trained for.
From these observations we conclude that our models, which are not trained to maximise the semantic meaning of the latent code, are so versatile, because they learn to encode the content of the image in the noise maps and to use the values of each pixel to further encode the color of the pixel, making them independent of the latent code in \stylegan, which is adjusted for one distribution only.
However, it would be interesting to investigate the latent projections of different encoders to learn more about the structure of the latent code in \stylegan.
We leave these experiments open for future work.

\begin{table*}
    \begin{center}
    \begin{tabular}{l | ccc | ccc | ccc | ccc}
    
    \multirowcell{3}{\small Training Dataset,\\ \small \stylegan Version,\\ \small Projection Target} & \multicolumn{12}{c}{Dataset and Metric for Evaluation} \\
                                    & \multicolumn{3}{c|}{FFHQ} & \multicolumn{3}{c|}{Church}   & \multicolumn{3}{c|}{Bedroom}  & \multicolumn{3}{c}{Cat}    \\
                                    & \fid  & \psnr & \ssim     & \fid  & \psnr & \ssim         & \fid  & \psnr & \ssim         & \fid  & \psnr & \ssim \\
    \hline\hline
    FFHQ, 2, $\mathcal{Z}$          &  5.87 & 25.19 & 0.91      &  8.28 & 19.95 & 0.84          &  4.96 & 23.45 & 0.89          &  3.90 & 22.65 & 0.87  \\
    FFHQ, 2, $\mathcal{W}^+$        &  0.37 & 29.48 & 0.96      &  1.21 & 23.99 & 0.93          &  0.48 & 27.96 & 0.96          &  0.83 & 25.95 & 0.93  \\
    \hline
    Church, 2, $\mathcal{Z}$        & 13.31 & 21.04 & 0.86      &  2.58 & 22.92 & 0.89          &  4.04 & 22.85 & 0.89          &  5.95 & 21.56 & 0.86  \\
    Church, 2, $\mathcal{W}^+$      &  2.06 & 24.18 & 0.92      &  0.19 & 30.12 & 0.96          &  0.40 & 27.73 & 0.96          &  0.96 & 25.40 & 0.92  \\
    \hline
    Bedroom, 2, $\mathcal{Z}$       &  8.05 & 24.07 & 0.90      &  3.05 & 22.45 & 0.88          &  2.28 & 26.17 & 0.92          &  2.85 & 23.70 & 0.89  \\
    Bedroom, 2, $\mathcal{W}^+$     &  0.75 & 26.75 & 0.94      &  0.62 & 26.69 & 0.95          &  0.17 & 31.64 & 0.97          &  0.53 & 27.28 & 0.93  \\
    \hline
    Cat, 2, $\mathcal{Z}$           & 10.72 & 24.31 & 0.91      &  4.12 & 22.23 & 0.87          &  4.62 & 24.87 & 0.90          &  3.62 & 24.42 & 0.90  \\
    Cat, 2, $\mathcal{W}^+$         &  0.92 & 28.06 & 0.95      &  0.48 & 28.15 & 0.96          &  0.25 & 30.55 & 0.97          &  0.48 & 28.60 & 0.94  \\
    \end{tabular}
    \end{center}
    \caption{
        The results of further reconstruction experiments.
        We use a \stylegan model for decoding, which was pre-trained on the LSUN cat dataset and not updated during the training of the encoder. 
        We train the decoder of our models on different datasets (FFHQ and LSUN datasets), \stylegan 2, and different projection targets ($\mathcal{Z},\mathcal{W}^+$) as shown in the first column.
        We evaluate each model on different datasets and report \ac{FID}, \ac{PSNR}, and \ac{SSIM}.
    }
    \label{tab:sup_cat_based_eval_results}
\end{table*}

\begin{table*}
    \begin{center}
    \begin{tabular}{l | ccc | ccc | ccc | ccc}
    
    \multirowcell{3}{\small Training Dataset,\\ \small \stylegan Version,\\ \small Projection Target} & \multicolumn{12}{c}{Dataset and Metric for Evaluation} \\
                                    & \multicolumn{3}{c|}{FFHQ} & \multicolumn{3}{c|}{Church}   & \multicolumn{3}{c|}{Bedroom}  & \multicolumn{3}{c}{Cat}    \\
                                    & \fid  & \psnr & \ssim     & \fid  & \psnr & \ssim         & \fid  & \psnr & \ssim         & \fid  & \psnr & \ssim \\
    \hline\hline
    FFHQ, 2, $\mathcal{Z}$          &  2.89 & 26.52 & 0.93      &  2.59 & 21.75 & 0.91          &  1.33 & 25.41 & 0.94          &  1.66 & 23.77 & 0.91  \\
    FFHQ, 2, $\mathcal{W}^+$        &  0.33 & 29.32 & 0.96      &  0.87 & 24.29 & 0.94          &  0.41 & 27.77 & 0.96          &  0.67 & 25.88 & 0.93  \\
    \hline
    Church, 2, $\mathcal{Z}$        &  6.75 & 21.82 & 0.89      &  0.85 & 25.69 & 0.92          &  1.92 & 23.77 & 0.92          &  2.52 & 22.49 & 0.89  \\
    Church, 2, $\mathcal{W}^+$      &  2.05 & 24.34 & 0.92      &  0.18 & 29.62 & 0.95          &  0.44 & 27.74 & 0.96          &  0.88 & 25.36 & 0.92  \\
    \hline
    Bedroom, 2, $\mathcal{Z}$       &  5.19 & 24.55 & 0.91      &  1.72 & 23.86 & 0.92          &  0.87 & 27.63 & 0.94          &  1.50 & 24.22 & 0.90  \\
    Bedroom, 2, $\mathcal{W}^+$     &  0.83 & 26.84 & 0.94      &  0.72 & 26.59 & 0.95          &  0.16 & 31.58 & 0.97          &  0.53 & 27.25 & 0.93  \\
    \hline
    Cat, 2, $\mathcal{Z}$           &  6.33 & 25.36 & 0.92      &  1.58 & 24.72 & 0.92          &  1.16 & 26.77 & 0.94          &  1.58 & 25.22 & 0.91  \\
    Cat, 2, $\mathcal{W}^+$         &  0.56 & 28.25 & 0.95      &  0.36 & 28.28 & 0.96          &  0.17 & 30.72 & 0.97          &  0.38 & 28.63 & 0.94  \\
    \end{tabular}
    \end{center}
    \caption{
        The results of further reconstruction experiments.
        Here, we use a \stylegan model for decoding, which was pre-trained on the LSUN church dataset and not updated during the training of the encoder. 
        We train the decoder of our models on different datasets (FFHQ and LSUN datasets), \stylegan 2, and different projection targets ($\mathcal{Z},\mathcal{W}^+$) as shown in the first column.
        We evaluate each model on different datasets and report \ac{FID}, \ac{PSNR}, and \ac{SSIM}.
    }
    \label{tab:sup_church_based_eval_results}
\end{table*}



\begin{table*}
    \begin{center}
    \begin{tabular}{l | ccc | ccc | ccc | ccc}
    
    \multirowcell{3}{\small Training Dataset,\\ \small \stylegan Version,\\ \small Projection Target} & \multicolumn{12}{c}{Dataset and Metric for Evaluation} \\
                                    & \multicolumn{3}{c|}{FFHQ} & \multicolumn{3}{c|}{Church}       & \multicolumn{3}{c|}{Bedroom}      & \multicolumn{3}{c}{Cat} \\
                                    & \fid  & \psnr & \ssim     & \fid  & \psnr & \ssim             & \fid  & \psnr & \ssim             & \fid  & \psnr & \ssim\\
    \hline\hline
    FFHQ, 2, $\mathcal{Z}$          & 24.27 & 15.35 & 0.69      & 38.62 & 12.76 & 0.60              & 41.53 & 14.15 & 0.66              & 29.47 & 13.77 & 0.63 \\
    FFHQ, 2, $\mathcal{W}^+$        &  9.73 & 22.16 & 0.85      & 22.53 & 17.33 & 0.74              & 20.35 & 19.09 & 0.80              & 15.20 & 18.92 & 0.79 \\
    \hline
    Church, 2, $\mathcal{Z}$        & 31.02 & 16.32 & 0.72      & 18.50 & 17.78 & 0.74              & 27.47 &17.48  & 0.75              & 23.33 & 16.35 & 0.71 \\
    Church, 2, $\mathcal{W}^+$      & 36.19 & 18.81 & 0.79      &  5.46 & 21.03 & 0.84              & 13.94 & 19.21 & 0.81              & 15.77 & 18.50 & 0.78 \\
    \hline
    Bedroom, 2, $\mathcal{Z}$       & 29.12 & 15.94 & 0.68      & 23.61 & 15.14 & 0.66              & 15.72 & 17.00 & 0.70              & 24.38 & 15.53 & 0.66 \\
    Bedroom, 2, $\mathcal{W}^+$     & 16.12 & 21.23 & 0.83      & 15.23 & 19.46 & 0.80              &  6.67 & 22.17 & 0.84              & 10.34 & 20.76 & 0.82 \\
    \hline
    Cat, 2, $\mathcal{Z}$           & 23.55 & 19.03 & 0.78      & 22.14 & 17.25 & 0.74              & 20.52 & 19.11 & 0.78              & 18.81 & 18.24 & 0.76 \\
    Cat, 2, $\mathcal{W}^+$         & 20.22 & 21.75 & 0.86      & 15.73 & 19.64 & 0.83              & 10.09 & 21.30 & 0.85              &  9.00 & 21.19 & 0.85 \\
    \end{tabular}
    \end{center}
    \caption{
        Image reconstruction results for models trained with our two network strategy.
        We can observe that these models are not as versatile as our other image reconstruction models that are not trained to maximise the semantic meaning of the latent code.
        However, the reconstruction quality is still high on the datasets they have been trained on.
    }
    \label{tab:sup_04_two_stem_quantitative}
\end{table*}

\end{document}